\pdfoutput=1

\documentclass[11pt]{article}

\usepackage[preprint]{acl}

\usepackage{times}
\usepackage{latexsym}

\usepackage[T1]{fontenc}

\usepackage[utf8]{inputenc}

\usepackage{microtype}

\usepackage{inconsolata}

\usepackage{graphicx}

\usepackage{booktabs}
\usepackage{multicol}
\usepackage{multirow}
\usepackage{makecell}
\usepackage{comment}
\usepackage{stackengine}
\usepackage{arydshln} 
\usepackage{amsmath}
\usepackage{amssymb}
\usepackage{amsthm}
\usepackage{circledsteps}
\usepackage[most]{tcolorbox}
\usepackage{listings}
\usepackage{xcolor}
\usepackage{enumitem}
\usepackage[capitalize,noabbrev]{cleveref} 

\title{RePPL: Recalibrating Perplexity by Uncertainty in Semantic Propagation and Language Generation for Explainable QA Hallucination Detection}

\author{
  \textbf{Yiming Huang\textsuperscript{1}\thanks{\quad Equal contribution.}},
  \textbf{Junyan Zhang\textsuperscript{1}\footnotemark[1]},
  \textbf{Zihao Wang\textsuperscript{2}},
  \textbf{Biquan Bie\textsuperscript{3}},
  \textbf{Yunzhong Qiu\textsuperscript{4}},
\\
  \textbf{Xuming Hu\textsuperscript{1,2}},
  \textbf{Yi R. (May) Fung \textsuperscript{2}},
  \textbf{Xinlei He \textsuperscript{1}\thanks{\quad Corresponding author.}}
\\
  \textsuperscript{1}The Hong Kong University of Science and Technology, Guangzhou, \\
  \textsuperscript{2}The Hong Kong University of Science and Technology, \\
  \textsuperscript{3}Independent Researcher,
  \textsuperscript{4}Tsinghua University
\\
\texttt{yhuang033@connect.hkust-gz.edu.cn}, \texttt{junyanzhang0317@gmail.com},   \\
\texttt{xinleihe@hkust-gz.edu.cn}
}

\begin{document}
\maketitle

\begin{abstract}
Large Language Models (LLMs) have become powerful, but hallucinations remain a vital obstacle to their trustworthy use.
Previous works improved the capability of hallucination detection by measuring uncertainty. But they can not explain the provenance behind why hallucinations occur, particularly in identifying which part of the inputs tends to trigger hallucinations.
Recent works on the prompt attack indicate that uncertainty exists in semantic propagation, where attention mechanisms gradually fuse local token information into high-level semantics across layers. Meanwhile, uncertainty also emerges in language generation, due to its probability-based selection of high-level semantics for sampled generations.
Based on that, we propose \textbf{RePPL} to \textbf{re}calibrate uncertainty measurement by these two aspects, which dispatches explainable uncertainty scores to each token and aggregates in \textbf{P}er\textbf{pl}exity-style Log-Average form as total score.
Experiments show that it achieves the best comprehensive detection performance across various QA datasets on advanced models (average AUC of 0.833), and it is capable of producing token-level uncertainty scores as explanations of hallucination.

\end{abstract}
\section{Introduction}
Since the milestone work ChatGPT~\citep{openai2023chatgpt} has revolutionized the fields of Language Generation, powerful Large Language Models (LLMs) are constantly updating and continuously shaping the text-driven automation in everyday applications like chatbots, document summarizers, and AI search engines.
Despite these successes, one of the fatal shortcomings of LLMs is hallucination, i.e., factually incorrect or contextually inconsistent outputs. Due to the urgent demand for trustworthiness in these LLM-based applications, detecting hallucination and figuring out how LLMs hallucinate is crucial to building robust AI systems.
\begin{figure}[!t]
    \centering
    \includegraphics[width=0.9\linewidth]{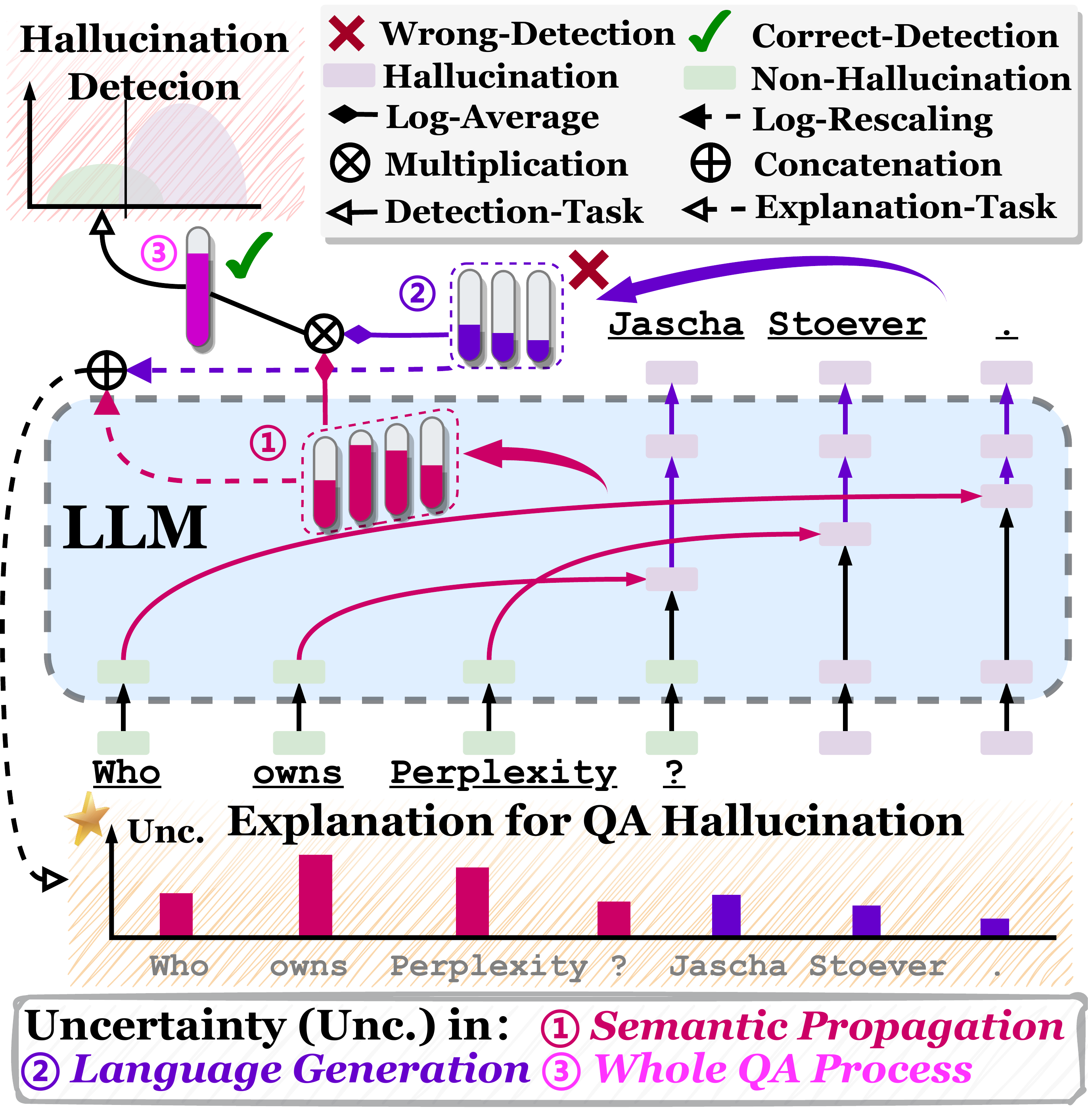}
    \caption{Illustration of our motivation. The LLM (Llama-3.1-8B-Instruct) generates a hallucinated answer ``Jascha Stoever'' to the question ``Who owns Perplexity?'', with high-confidence (low-uncertainty) logits. This makes the detection hard for existing methods using logits. While our method judges it correctly with the explanation, by balancing uncertainty in semantic propagation and language generation.}
    \label{fig:mov}
\end{figure}

To tackle question answering (QA) hallucination in LLM, prior works~\citep{Perplexity, EnergyScore, P(True), LNPE, SemanticEnt, EigenScore} identify the uncertainty of generations as the key point. These methods can be divided into two groups: logits-based and sampling-based methods.
The logits-based methods~\citep{Perplexity, EnergyScore, P(True)} utilize the confidence of each token's generation to accumulate uncertainty, while the sampling-based~\citep{LNPE, SemanticEnt, EigenScore} methods measure the discrepancy between sampled generations as uncertainty.
However, these detection methods cannot explain why LLMs hallucinate and which part of the inputs mainly triggers the hallucination.

Our motivation stems from two observations. Firstly, recent works~\citep{LLMFoolSelf, AutoDAN} about LLMs' prompt attack applied to different parts of inputs can easily trigger various LLMs' untrustworthy output, including hallucinations. This suggests that uncertainty arises in semantic propagation, where attention gradually fuses local token information into high-level semantics across layers.
Secondly, the confidences of generations represented by logits often vary between different tokens and different generations. It implies that probability-based token generation selectively discards the remaining high-level semantics, bringing uncertainty in language generation.
As~\Cref{fig:mov} shows, we aim to combine these two aspects for better hallucination detection.

From this viewpoint, we propose \textbf{RePPL} to \textbf{re}calibrate uncertainty by these two aspects with \textbf{P}er\textbf{pl}exity-style aggregation.
For the intrinsic semantic propagation, we utilize the attribution-based explainability method~\citep{Rollout, GenAtt} to compute token-level uncertainty (inverse of confidence) by measuring token attribution discrepancies across sampled generations. 
For the extrinsic language generation, we normalize token confidence by the average generation length of sampling to make the confidence more distinct. 
As~\Cref{fig:mov} depicts, uncertainty (or confidence) scores correspond to each token as the explanation in these two aspects. And we use Perplexity-style~\citep{Perplexity} aggregation that firstly rescales confidence into log form, then averages them all. We name these two total scores as \textbf{InnerPPL} and \textbf{OuterPPL}.
Through multiplying them, we derive the recalibrated hallucination score \textbf{RePPL} with hierarchical distinctness from the above two aspects.
Experiments show that our method comprehensively outperforms baselines across QA datasets, achieving an average AUC of 0.833.
It also faithfully explains hallucination by highlighting the uncertainty-inducing tokens.

To summarize, the contribution of this work can be concluded as the following threefold:
\begin{enumerate}
    \item We address the novel view of balancing uncertainty in semantic propagation and language generation, which brings recalibrated uncertainty with token-correspondence and hierarchical distinctness to propose our innovative hallucination detection score \textbf{RePPL}.
    \item Our proposed \textbf{RePPL} achieves the leading performance in hallucination detection tasks across various datasets and models. And it is also robust under different hyperparameter settings and sampling options.
    \item \textbf{RePPL}'s uncertainty scores are pioneering in explaining the hallucination triggered by both input prompts and output contents. It provides a faithful token-level explanation by indicating uncertainty-inducing tokens. 

\end{enumerate}
\section{Related Works and Background}
\subsection{Uncertainty in Hallucination Detection}
Uncertainty measurement~\citep{UncSurvey, WeightUnc} is a classic research topic in the wide machine learning domain. It is highly related to out-of-distribution detection, which is inherently correlated to hallucination detection~\citep{zhang2024r, EstUnc,he2025mmboundary}.
For non-parameterized methods, uncertainty-based hallucination detection can be mainly sorted into two kinds.
The first is the logits-based, such as widely used Perplexity~\citep{Perplexity} and typical Verbalize~\citep{P(True)} a.k.a. P(True), these methods exert confidence within logits to figure up uncertainty in a single generation.
The second is the sampling-based, for instance, Semantic Entropy~\citep{SemanticEnt} and EigenScore~\citep{EigenScore}.
These methods take the discrepancy between different generations as uncertainty.
Despite the effectiveness of these methods, they only focus on uncertainty in generation.
It hinders them from exploiting the semantic propagation about how input transforms into output, for better performance and potential explainability.

For parameterized methods, we also notice the powerful probe-based method~\citep{LLMmoreThan, liu2024universal, FacLens}.
But they have two main shortcomings:
1. Regarding binary classification, they need adjustments and extra implementations with task-specific losses or hyperparameters for peak performance on abundant training data.
2. As for the hallucination degree comparator, they are unsuitable as a zero-shot hallucination degree indicator for scenarios like tracing and comparing hallucination risks of the LLM in the pretraining stage.
We hence consider them more cost-effective for target LLMs and additionally compare them in~\Cref{ap:rouge} and~\ref{ap:probe} with our method.

\subsection{Attribution Explanation}
In the explainable AI (XAI) domain, attribution is an enduringly appealing topic as it indicates the salient part of the model's inputs to the model's prediction.
For model-specific ones, Rollout~\citep{Rollout}, GenAtt~\citep{GenAtt}, and AttnLRP~\citep{AttnLRP} improve the attribution ability by utilizing attention maps among different layers of transformers.
For model-agnostic ones, like LIME-style~\citep{LIME} and Shapley-value-based~\citep{SHAP1, SHAP2}, they need sufficient input perturbation for measuring the importance of absent tokens.
Because model-specific methods need expensive gradients and model-agnostic methods need too many throughputs to guarantee their precision, they are intractable in explaining real-time hallucination detection.
Therefore, we adopt simple maximum, average, and Rollout to pool attention maps as attribution across layers and heads.
These attribution methods commonly generate an attribution map, and the row of this map represents the token-to-token contribution importance (scores).
At the instance level, we attribute the uncertainty score expressing token-wise propensity to induce hallucinations rather than the importance score by the above methods, as the explanation of hallucination.
At the macro-level, it is also useful to explore the underlying factors of the hallucination, like prior works~\citep{zhang2025lawknowledgeovershadowingunderstanding, DoIE, MechHallu, LLMmoreThan}.
\label{sec:2.2}

\section{Methodologies}
\begin{figure*}[!t]
    \centering
    \includegraphics[width=1.0\linewidth]{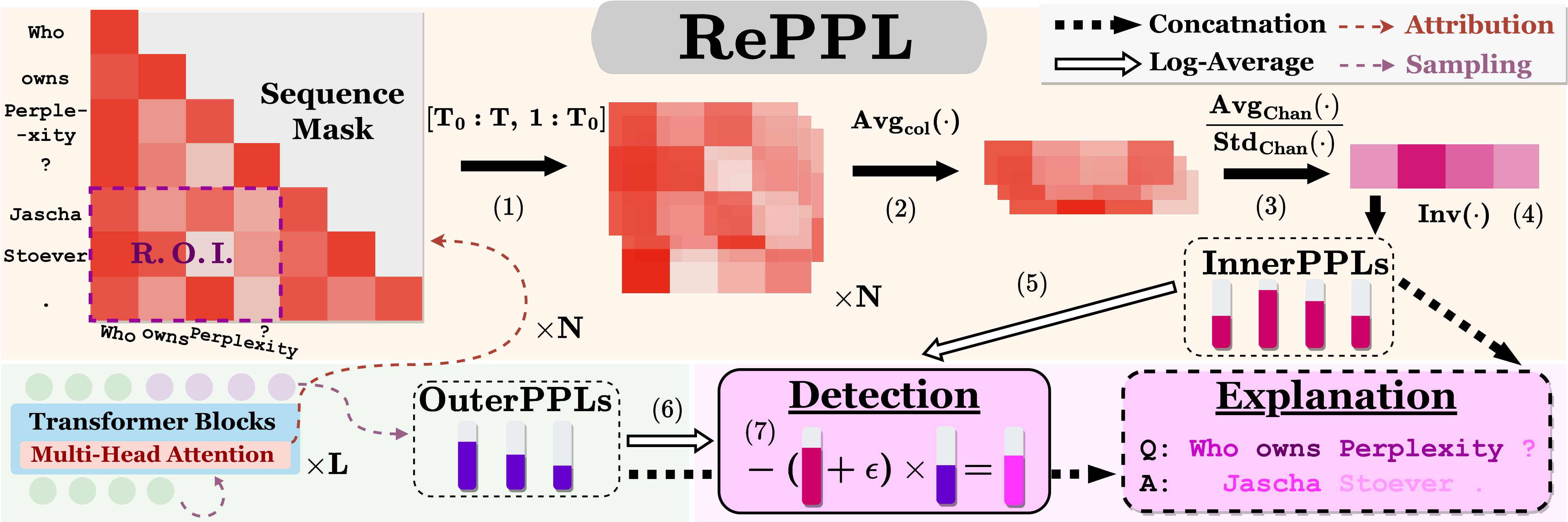}
    \caption{Detailed illustration of \textbf{RePPL} computing process. $(1)\sim(7)$ correspond to the same equations in paper. `\textbf{InnerPPLs}' and `\textbf{OuterPPLs}' with plural form to indicate token-wise confidence scores.}
    \label{fig:method}
\end{figure*}
\subsection{Attention-based Attribution}
As discussed above, we choose the following three attention pooling strategies.
The first is maximum pooling, abbreviated as \textbf{MaxPool}. Given attention maps of $\mathbf{t}$ tokens $\mathbf{A^{(l)}} \in \mathbb{R}^{h\times t\times t}$ across different attention heads $\mathbf{h}$ of attention in layer $\mathbf{l}$ in $\mathbf{L}$ layers transformer-based LLM. We take the final attribution $\mathbf{R} \in \mathbb{R}^{t \times t}$ as:
\begin{align}
     \mathbf{R} & = \operatorname{MaxPool(\mathcal{A})}
     = \operatorname{Max_l}(\operatorname{Max_h}(\mathcal{A}))\ , \notag
\end{align}
here, $\mathcal{A} = \mathbf{[A_1, A_2, .. A_L]}$ is the attention maps concatenation along the layer dimension and $\operatorname{Max_d}$ is the maximum selection among values along $\mathbf{d}$ dimension. Similarly, the average pooling \textbf{AvgPool} is ($\operatorname{Avg_d}$ is the average operation):
\begin{align}
    \mathbf{R} & = \operatorname{AvgPool(\mathcal{A})} = \operatorname{Avg_l}(\operatorname{Avg_h}(\mathcal{A})) \ . \notag
\end{align}
As for Rollout~\citep{Rollout} pooling \textbf{RollPool}, it layer-wisely accumulates relevance from equal single-head attention $\mathbf{\bar{A}}$ as attribution, meanwhile, it also considers the relevance in the residual part. We adopt same enhancement in GenAtt~\citep{GenAtt} as follows:
\begin{align}
    & \mathbf{R^{(l)}} \leftarrow  \mathbf{R^{(l-1)}} +  \mathbf{R^{(l-1)}} \cdot \mathbf{\bar{A}_l} \ , \notag \\ & \mathbf{\bar{A}_l} = \operatorname{Norm_{row}}(\operatorname{Avg_h}(\mathbf{A^{+}_l}) - \mathbb{I}) / 2 + \mathbb{I}/2 \ ,\notag
\end{align}
therein, it begin with the initialization $\mathbf{R^{(0)}} := \mathbb{I}$, $\operatorname{Norm_{row}}$ is the normalization ($\mathbf{p}=1$) across horizonal rows of matrix, $\mathbb{I}$ is the indentical matrix and we count the $\mathbf{R^{(L)}}$ end up at layer $\mathbf{L}$ as the final attribution.
In our application context, these attributions prepare the granular token-to-token importance scores for both detailed and explainable semantic propagation measurement, and they are more efficient than others.
\label{sec:3.1}
\subsection{Uncertainty Recalibration}
\paragraph{Overview.} The final uncertainty score is recalibrated by multiplying the inner semantic propagation uncertainty derived from attribution variance across samples, and the outer language generation uncertainty based on a modified aggregation of token-level confidence.

\begin{figure}[!htbp]
    \centering
    \includegraphics[width=1.0\linewidth]{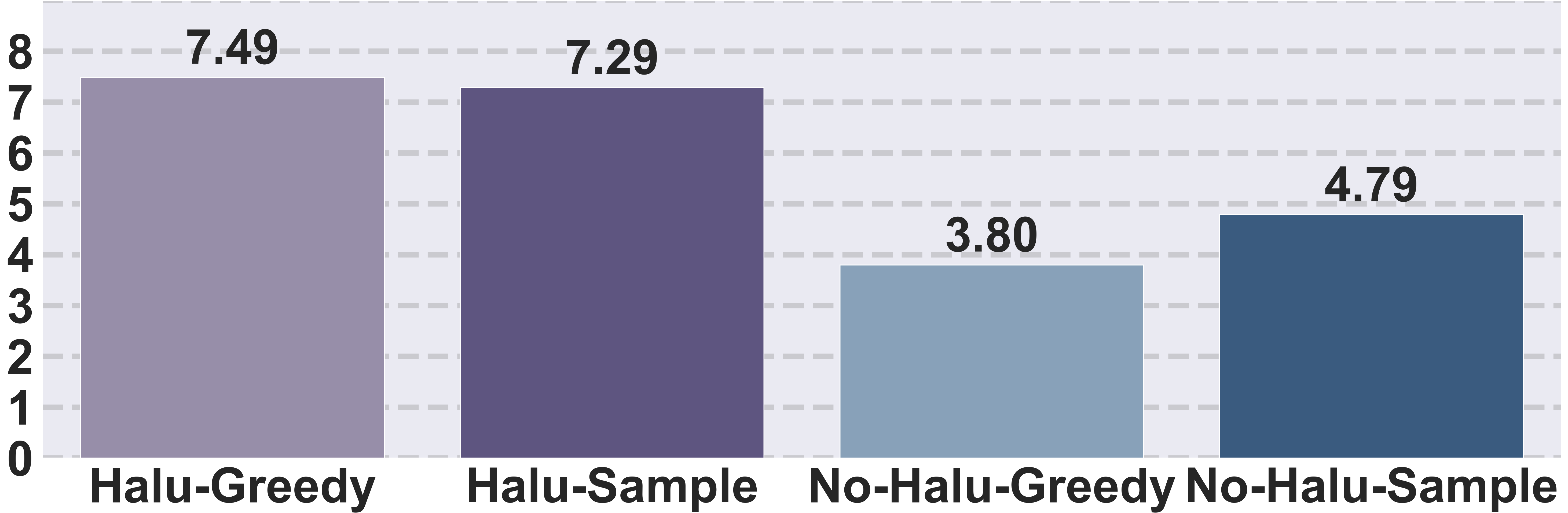}
    \caption{Length of greedy (`Greedy') outputs and average length of sampled generations (`Sample') of 1000 examples in TriviaQA. `Halu' is hallucinated ones, and `No-Halu' is the not hallucinated ones.}
    \label{fig:gs}
\end{figure}

\paragraph{InnerPPL.} For inner semantic propagation, providing the abovementioned attribution matrix $\mathbf{R}$ (also row-normalized) after LLM's inference (detailed in~\Cref{fig:method}), we convert importance to the uncertainty score of each token and aggregate them as the following steps.
Firstly, we sample $\mathbf{N}$ times from the LLM, which empowers uncertainty to appear between the attribution matrices $\{\mathbf{R_1}, \mathbf{R_2}, ...,\mathbf{R_N}\}, \mathbf{R_n} \in \mathbb{R}^{T_n}$ of different generations, $\mathbf{T_n}$ is the total length of input tokens added up with output tokens and $\mathbf{T_0}$ in the length of input tokens.
In these attribution matrices, the region of interest (\textbf{R.O.I.}) is the columns from the first column to the $\mathbf{T_0}$ column in rows from the $\mathbf{T_0}$ row to the $\mathbf{T_n}$ row,
because they represent each input token's semantic contribution in the propagation process of each output token.
These regions reflect how the input token propagates its own semantics into high-level semantics, which are selected by generated tokens. That is:
\begin{align}
    \mathcal{R}_{roi} = \{ \mathbf{R_{n, [T_0:T_n, \ 1:T_0]}}| n \in \{1, 2,..., N\}\}.
\end{align}
Here, $\mathbf{[a:b, c:d]}$ is the continuous submatrix selection operation which selects the submatrix of the a-th row to b-th row and c-th column to d-th column of .
Thus, the uncertainty of semantic propagation can be derived by measuring the discrepancy of these regions in different sampling channels.
Then, we average different columns across the rows of each attribution submatrix ($\operatorname{col}$ refers to columns dimension):
\begin{align}
    \bar{\mathcal{R}}_{roi} = \operatorname{Avg_{col}}(\mathcal{R}_{roi}) \in \mathbb{R}^{N\times T_0}.
\end{align}
These attribution submatrices are condensed into vectors with $\mathbf{T_0}$ length and $\mathbf{N}$ sampling channels.
They express the average semantic propagation on all generated tokens, which also gets rid of the different generation lengths.
To derive uncertainty in sampling channels, we choose the coefficient of variation that is scale-/unit-independent for undistorted uncertainty measurement.
That is the following equation ($\operatorname{chan}$ refers to channels dimension):
\begin{align}
    \mathbf{r} = \frac{\operatorname{Std_{chan}}(\bar{\mathcal{R}}_{roi})}{\operatorname{Avg_{chan}}(\bar{\mathcal{R}}_{roi})},
\end{align}
it directly uses the standard deviation ($\operatorname{Std}$) divided by the average. The coefficient of variation vector $\mathbf{r}$ quantifies the uncertainty of different input tokens.
Next, we use the inverse function to transform this coefficient of variation vector to pseudo-confidences similar to logit-based probabilities:
\begin{align}
    \hat{\mathbf{p}} \in {[0, 1]}^{T_0},\ \hat{p}_i = \operatorname{Inv}(r_i) =\frac{1}{1 + r_i^\alpha}. \label{eq4}
\end{align}
Here, $\alpha$ is the hyperparameter that controls uncertainty scaling.
Subsequently, similar to the Perplexity~\cite{Perplexity}, we adopt log rescaling and length-as-divisor average of $\hat{\mathbf{p}}$ to calculate the total score in inner side semantic propagation:
\begin{align}
    \textbf{InnerPPL} = - \frac{1}{T_0}\sum_{i=1}^{T_0} log(\mathbf{\hat{p}}_i) \ .
\end{align}
The $log$ enlarges the significant part of the uncertainty, thus more validly takes it into account for hallucination.
\paragraph{OuterPPL.} For the outer part of language generation, given the $\mathbf{g}$-th generated token confidence $\mathbf{p}_g$ and generation length $S_g$ at the greedy decoding stage, we compute the overall generation uncertainty as:
\begin{align}
    \textbf{OuterPPL}& = -\frac{1}{\bar{S}}\sum^{S_g}_{\mathbf{g}=1} log(\mathbf{p_g}) \ ,
\end{align}
$\bar{S}$ is the average length of $\mathbf{N}$ sampled generations. The only difference between our \textbf{OuterPPL} and the original PPL is different divisors.
The rationale behind this divisor is that we observe (as shown in~\Cref{fig:gs}) that the generation length of the greedy decoding is basically shorter than sampling generations when LLM is certain.
On the contrary, if LLM tends to hallucinate, the greedy path is roughly equal to most sampling paths.
Thus, this divisor plays the role of a reward or penalty factor to make outer uncertainty more discriminative.
\paragraph{RePPL.} Finally, our hallucination detection score is derived from the multiplication of these two scores, which means recalibration of uncertainty from the two aspects. It is:
\begin{align}
     \textbf{RePPL}= - (\textbf{InnerPPL} + \epsilon) \times \textbf{OuterPPL}. \label{eq:7}
\end{align}
In the above equation, $\epsilon$ is the preset bias term, which is the default value to express correct uncertainty in case \textbf{InnerPPL} is 0 but \textbf{OuterPPL} is still large. In general, we choose $\frac{1}{10} \sim \frac{1}{5}$ of average \textbf{InnerPPL} value for $\epsilon$.
Since the multiplication form of our hallucination score comprehensively considers uncertainty in semantic propagation and language generation, it is capable of hierarchical division when \textbf{InnerPPL} is large but \textbf{OuterPPL} is small, or vice versa. This trait equipped our method more layered distinctness for better detection.
As for the uncertainty (or confidence) scores that granularly express token-wise contribution to hallucination, we collect them from the above computing process as an explanation of hallucination.
These token-by-token scores attribute uncertainty rather than importance, thus figure out which input tokens trigger hallucination and which output tokens are more likely to be the hallucination.
\label{sec:3.2}
\section{Evaluation}
\subsection{Experimental Setup}
\begin{table*}[!t]
\centering
\small
\resizebox{1\textwidth}{!}
{
\setlength{\tabcolsep}{1.0pt}
\renewcommand{\arraystretch}{1.0}
\begin{tabular}{llcccccccccccccccccccc}
\toprule
\multirow{2}{*}{\textbf{Models}} & \textbf{Datasets}
& \multicolumn{4}{c}{\textbf{CoQA}}
& \multicolumn{4}{c}{\textbf{TriviaQA}}
& \multicolumn{4}{c}{\textbf{NQ}}
& \multicolumn{4}{c}{\textbf{SQuAD}}
& \multicolumn{4}{c}{\textbf{ALL}} \\
\cmidrule(lr){3-6} \cmidrule(lr){7-10} \cmidrule(lr){11-14} \cmidrule(lr){15-18} \cmidrule(lr){19-22}
& \textbf{Methods} & AUC & Acc & Corr & PRR
& AUC & Acc & Corr & PRR
& AUC & Acc & Corr & PRR
& AUC & Acc & Corr & PRR 
& AUC & Acc & Corr & PRR \\
\midrule
\multirow{8}{*}{\makecell[c]{\textbf{LLaMA-3.1-} \\ \textbf{8B-Instruct}}}
& Perplexity & 66.6 & 61.7 & 30.0 & 41.7 & 78.2 & 71.1 & 51.0 & 64.3 & 77.5 & 67.5 & 38.9 & 46.1 & 67.4 & 64.9 & 31.1 & 31.3 & 73.9 & 67.5 & 43.8 & 52.8 \\
& Energy & 64.4 & 59.6 & 26.5 & 39.2 & 67.6 & 62.6 & 33.7 & 42.7 & 66.3 & 65.3 & 26.2 & 28.3 & 69.4 & 64.1 & 33.4 & 46.5 & 69.1 & 63.4 & 35.8 & 48.3 \\
& P(True) & 64.0 & 60.2 & 25.8 & 35.1 & 75.8 & 69.6 & 48.7 & 63.0 & 69.1 & 67.0 & 33.2 & 38.6 & 64.4 & 60.4 & 25.7 & 39.7 & 71.5 & 65.8 & 40.9 & 53.0 \\
& LNPE & 72.1 & 65.7 & 39.9 & 50.3 & 80.2 & 73.0 & 55.2 & 67.6 & 78.1 & 68.1 & 42.8 & 48.4 & 69.7 & 67.2 & 35.1 & 34.4 & 76.8 & 70.1 & 49.5 & 58.5 \\
& Semantic Entropy & 74.5 & 70.6 & 43.2 & 54.3 & 73.6 & 69.4 & 43.9 & 55.9 & 67.9 & 69.0 & 24.3 & 30.1 & 66.7 & 68.6 & 31.1 & 39.8 & 73.9 & 70.4 & 44.0 & 54.0 \\
& EigenScore & 77.2 & 70.6 & 49.3 & 62.8 & 80.5 & 74.3 & 55.3 & 67.7 & 74.9 & \textbf{71.6} & 42.4 & 46.8 & \textbf{76.8} & 70.9 & \textbf{47.4} & \textbf{61.1} & 78.2 & 71.8 & 51.8 & 63.7 \\
& AttnScore & 75.0 & 69.4 & 45.0 & 48.7 & 76.0 & 69.2 & 44.4 & 50.0 & 67.9 & 62.0 & 26.6 & 28.1 & 64.5 & 61.7 & 26.3 & 30.3 & 62.6 & 60.2 & 21.8 & 31.2 \\
& \textbf{RePPL (Ours)} & \textbf{85.2} & \textbf{77.4} & \textbf{61.8} & \textbf{70.7} & \textbf{86.0} & \textbf{78.6} & \textbf{63.2} & \textbf{73.9} & \textbf{79.3} & 69.9 & \textbf{47.4} & \textbf{51.9} & 76.0 & \textbf{71.7} & 45.1 & 40.6 & \textbf{83.6} & \textbf{76.3} & \textbf{60.0} & \textbf{66.4} \\
\midrule
\multirow{8}{*}{\makecell[c]{\textbf{Qwen-2.5-} \\ \textbf{7B-Instruct}}}
& Perplexity & 74.9 & 71.7 & 43.3 & 58.0 & 83.6 & 75.9 & 60.5 & 71.4 & 76.8 & 73.3 & 39.0 & 42.4 & 75.5 & 69.8 & 43.5 & 52.3 & 80.6 & 74.3 & 55.3 & 65.5 \\
& Energy & 53.0 & 51.8 & 9.7 & 17.1 & 76.3 & 69.3 & 50.9 & 58.6 & 64.8 & 57.1 & 35.6 & 33.8 & 59.1 & 57.7 & 14.3 & 19.9 & 66.1 & 61.0 & 33.8 & 39.4 \\
& P(True) & 54.7 & 53.3 & 10.3 & 11.1 & 79.7 & 74.3 & 55.2 & 65.8 & 67.5 & 68.8 & 22.3 & 27.6 & 56.8 & 57.0 & 12.3 & 13.7 & 70.6 & 65.3 & 39.8 & 46.2 \\
& LNPE & 76.8 & 71.9 & 46.8 & 58.8 & 83.6 & 76.1 & 61.2 & 72.7 & 77.4 & 69.1 & 42.4 & 42.8 & 76.1 & 70.5 & 44.6 & 52.7 & 81.2 & 74.3 & 56.9 & 66.3 \\
& Semantic Entropy & 71.3 & 69.2 & 39.3 & 42.6 & 76.7 & 76.2 & 48.3 & 57.6 & 72.1 & \textbf{76.9} & 24.5 & 28.8 & 60.8 & 54.6 & 19.7 & 25.0 & 74.5 & 72.8 & 46.0 & 50.5 \\
& EigenScore & 83.0 & 76.2 & 58.4 & 61.3 & 82.5 & 75.2 & 60.2 & 66.4 & 79.5 & 69.7 & \textbf{49.2} & \textbf{47.1} & 77.3 & \textbf{73.4} & 46.9 & 54.1 & 82.5 & 75.2 & 59.8 & 65.0 \\
& AttnScore & 70.7 & 65.0 & 34.8 & 44.1 & 60.4 & 57.3 & 18.6 & 18.1 & 63.6 & 58.7 & 12.1 & 12.0 & 69.9 & 65.8 & 35.2 & 43.0 & 57.9 & 62.0 & 16.1 & 27.4 \\
& \textbf{RePPL (Ours)} & \textbf{83.5} & \textbf{77.0} & \textbf{58.4} & \textbf{67.2} & \textbf{84.6} & \textbf{76.3} & \textbf{62.4} & \textbf{73.3} & \textbf{80.0} & 73.6 & 45.8 & 46.5 & \textbf{78.1} & 71.9 & \textbf{47.8} & \textbf{54.9} & \textbf{83.7} & \textbf{75.9} & \textbf{60.9} & \textbf{68.9} \\
\midrule
\multirow{8}{*}{\makecell[c]{\textbf{Qwen-2.5-} \\ \textbf{14B-Instruct}}}
& Perplexity & 74.5 & 73.6 & 39.2 & 43.8 & 81.4 & 75.3 & 56.4 & 67.1 & 76.7 & 70.7 & 44.2 & 46.6 & 75.7 & 71.5 & 45.2 & 48.6 & 78.0 & 73.4 & 49.4 & 56.3 \\
& Energy & 51.3 & 48.9 & 18.3 & 18.1 & 62.5 & 58.6 & 26.5 & 29.4 & 58.0 & 56.8 & 28.9 & 29.0 & 60.8 & 58.6 & 18.4 & 21.8 & 52.8 & 50.2 & 11.4 & 14.0 \\
& P(True) & 58.9 & 54.4 & 20.9 & 22.1 & 70.1 & 68.1 & 41.7 & 42.2 & 63.8 & 66.6 & 23.9 & 22.9 & 53.2 & 52.9 & 8.3 & 0.1 & 60.0 & 57.4 & 23.3 & 19.4 \\
& LNPE & 76.1 & 75.8 & 41.3 & 46.8 & 82.4 & 75.5 & 58.6 & 69.3 & 77.2 & \textbf{72.2} & 46.8 & 49.5 & 76.9 & 71.4 & 47.4 & 51.1 & 79.1 & 74.2 & 51.4 & 57.9 \\
& Semantic Entropy & 66.2 & 53.6 & 27.9 & 22.4 & 73.9 & 72.9 & 47.2 & 49.6 & 71.7 & 71.4 & 31.9 & 34.1 & 63.2 & 59.7 & 26.1 & 24.0 & 69.3 & 64.1 & 37.7 & 34.7 \\
& EigenScore & \textbf{82.5} & 72.2 & \textbf{49.7} & \textbf{49.8} & 81.8 & 75.5 & 59.1 & 65.1 & 77.7 & 70.8 & \textbf{51.7} & 46.4 & 80.4 & 73.0 & 53.5 & 54.0 & 81.6 & 73.2 & \textbf{56.7} & 56.4 \\
& AttnScore & 51.9 & 51.9 & 7.6 & 8.6 & 64.5 & 61.0 & 27.5 & 30.6 & 64.5 & 59.4 & 25.1 & 20.6 & 54.5 & 53.9 & 9.3 & 1.4 & 64.4 & 64.1 & 24.2 & 30.7 \\
& \textbf{RePPL (Ours)} & 79.7 & \textbf{76.6} & 43.0 & 47.9 & \textbf{84.8} & \textbf{76.8} & \textbf{62.2} & \textbf{71.9} & \textbf{80.0} & 71.7 & 51.4 & \textbf{50.1} & \textbf{81.3} & \textbf{74.4} & \textbf{54.7} & \textbf{56.8} & \textbf{82.5} & \textbf{75.7} & 56.0 & \textbf{60.4} \\
\bottomrule
\end{tabular}
}
\caption{Evaluation results of different methods on various datasets using AUC, accuracy at threshold of max G-Mean (Acc), and Spearman Correlation (Corr) metrics. Embedding (all-MiniLM-L6-v2) similarity is the correctness measure. We also present performance on the 4 datasets' merge in the ``\textbf{ALL}'' columns. The highest values are highlighted in bold, and all values are in percentages.}
\label{tab:main_results}
\end{table*}
\paragraph{Datasets.}
Question answering (QA) is the most common scenario for LLM-based applications.
We choose four widely used QA datasets to evaluate our proposed hallucination score.
Two of them are closed-book QA datasets, i.e., TriviaQA~\citep{TriviaQA} and Natural Questions (NQ)~\citep{NQ}.
TriviaQA is a dataset that assembles trivia knowledge questions with precise answers,
we select 9960 QA pairs in the validation split of \textit{rc.noncontext} subset after deduplication.
NQ is a dataset composed of Google search-based QA pairs. We choose the 3610 QA pairs in the validation split for evaluation, same as~\citet{EigenScore}. The other two are open-book QA datasets, i.e., CoQA~\citep{COQA} and SQuAD~\citep{SQuAD}. CoQA is a dataset about conversational QA, we utilize the development split of CoQA with 7983 QA pairs for our experiment. As for SQuAD, it is a reading comprehension dataset, and we take the development-\textit{v2.0} split with 5928 QA pairs (with filter condition \textit{is\_impossible=True}). All of the above subset selections are the same as the implementation of prior work of~\citet{EigenScore}.
\paragraph{Target Models.}
We choose three representative open-source models {Llama-3.1-8B-Instruct}~\citep{grattafiori2024llama}, {Qwen2.5-7B-Instruct}, and {Qwen2.5-14B-Instruct}~\citep{yang2024qwen2} for experiments.
Currently, they are widely used in diversified applications, and they are popular in various leadboards.
Since they are high-performance instruction-finetuned models, we believe their error can represent hallucination issues well in general application scenarios.
\paragraph{Inference Settings.}
On account of the instruction finetuned models, we equip them with a normal chat template described in~\Cref{ap:set}.
For sampling-based methods, including ours, in line with prior works~\citep{SemanticEnt, EigenScore}, we set the number of sampled generations $\mathbf{N} = 10$, temperature $\mathcal{T} = 1.0$, sampling hyperparameters $\text{Top-K} = 50$ and $\text{Top-P} = 0.99$ to guarantee the randomness needed by these methods.
These inference settings are both fair for comparison and practical for application.
\paragraph{Correctness Measure.}
Consistent with previous work~\citep{EigenScore}, we take the output in greedy decoding as LLM's canonical response to judge whether it is hallucinated.
We follow~\citet{HaluScope} and~\citet{EigenScore} work to use embedding similarity as a correctness measure for ground truth labeling.
The sentence embedding of the LLM's response and the standard answer of the QA pairs are extracted by the model all-MiniLM-L6-v2{~\citep{reimers2019sentence}}.
We treat embedding similarity between the LLM's response and the standard answers lower than 0.9 as hallucination.
Besides that, we also supplement the experiment using ROUGE-L (f-measure)~\citep{ROUGE} as the correctness measure in~\Cref{ap:rouge}.
\paragraph{Evaluation Metrics.}
To comprehensively evaluate the performance of hallucination detection scores, similar to~\citet{EigenScore}, we examine them by three metrics.
The first is the widely acknowledged area under the receiver operator characteristic curve (AUC).
The second is accuracy (Acc) at the classification threshold of deriving the maximum \textbf{G-Mean} value.
G-Mean is defined as $\textbf{G-Mean} = \sqrt{\text{TPR} \times (1- \text{FPR})}$, where $\text{TPR}$ is true positive rate and $\text{FPR}$ is false positive rate.
This accuracy is well recognized when evaluating classification performance in imbalanced situations.
Thirdly, we select Spearman Correlation (Corr) to fairly compare the correlation between hallucination scores and ground truth labels.
Lastly, we also use the PRR metric in recent work of~\citet{PRR}, which is a well-established metric for evaluating uncertainty quantification methods. 
For these metrics, higher scores indicate better performance. These metrics are detailed in~\Cref{ap:metric}.
\paragraph{Baselines.}
We select six baselines for comparison.
The logits-based methods include popular \textbf{Perplexity}~\citep{Perplexity}, commonly applied OOD detection score \textbf{Energy}~\citep{EnergyScore}, and \textbf{P(True)}~\citep{P(True)}.
The sampling-based methods contain Length-normalized Predictive Entropy (\textbf{LNPE})~\citep{LNPE} , broadly acknowledged \textbf{Semantic Entropy}~\citep{SemanticEnt} and \textbf{EigenScore}~\citep{EigenScore}. Besides, we compare the internal uncertainty method \textbf{AttnScore}~\citep{LLM_Check}.
These baselines are detailed in~\Cref{ap:baseline}. We also compare the parameterized probe-based methods SAPLMA~\citep{SAPLMA}, Probe-LR~\citep{ProbeLR}, and Probe-MM~\citep{ProbeMM} in~\Cref{ap:probe}. 
\label{sec:4.1}

\begin{figure*}[!ht]
    \centering
    \includegraphics[width=1.0\linewidth]{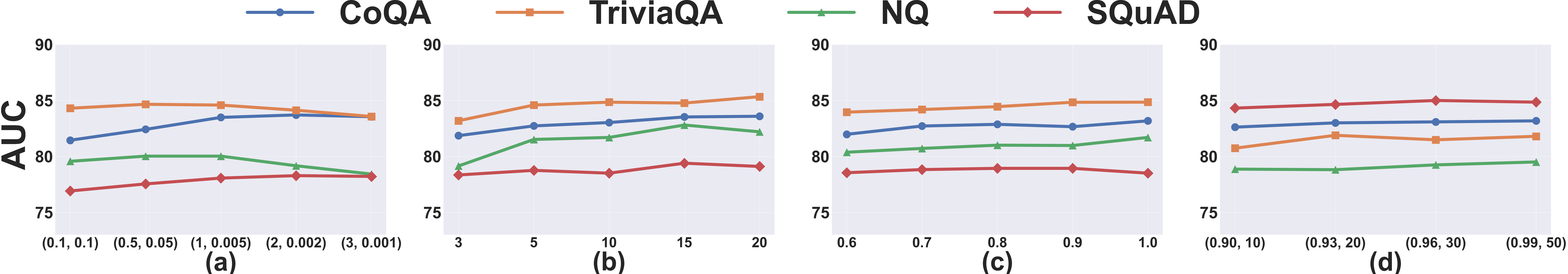}
    \caption{Performance under different hyperparameters. They are (a) our rescaling hypermeter group $(\alpha, \epsilon)$, (b) number of Generation $\mathbf{N}$, (c) temperature $\mathcal{T}$, (d) sampling hyperparameter group $(\text{Top-P}$, $\text{Top-K})$.}
    \label{fig:plot}
\end{figure*}
\subsection{Main Results}
We select \textbf{AvgPool}, $\alpha=1.0$, and $\epsilon=0.005$ for our method.
\Cref{tab:main_results} reports the main results about the performance comparison of our \textbf{RePPL} and other baseline methods.
From these results, we find that:
(1) Our method achieves strong performance on hallucination detection. We reach the average AUC of \textbf{0.813} and \textbf{74.65\%} accuracy at maximum G-Mean value on all individual datasets and all models. In addition, we also achieve an average AUC of \textbf{0.833} across 4 datasets (on the merge of 4 datasets).
(2) Our method outperforms baselines, especially the strong baseline LNPE and EigenScore.
Out of \textbf{60} evaluated metrics across datasets and models, our method ranks first in \textbf{46}, demonstrating its overall superiority.
Compared to EigenScore, which ranks second and performs an average AUC of 0.795 \& 72.78\% average accuracy,
Our method takes the leading position of detection performance by 2\% improvement.
And we also outperform in the merge of all datasets.
(3) Our method exhibits high generalization across different datasets.
Our method basically guarantees the AUC higher than 0.75, and the accuracy higher than 70\%.
(4) Our method also demonstrates good adaptability among different models, and it can scale to larger models like \textbf{Qwen2.5-14B-Instruct}.
Therefore, \textbf{RePPL} owns strong hallucination detection performance. And it firmly supports the foundation of its explainability. Although our method brings a certain light increase to runtime compared to other sampling-based methods (see in~\Cref{ap:runtime}) and fails to outperform the probe-based method (\Cref{ap:probe}), this non-parameteric zero-shot hallucination indicator unlocks rich explainability for analyzing hallucination patterns at micro-level (\Cref{sec:5.2}) \& macro-level (\Cref{ap:relation}) as a trade-off.

\begin{figure}[!htbp]
    \centering
    \includegraphics[width=1.0\linewidth]{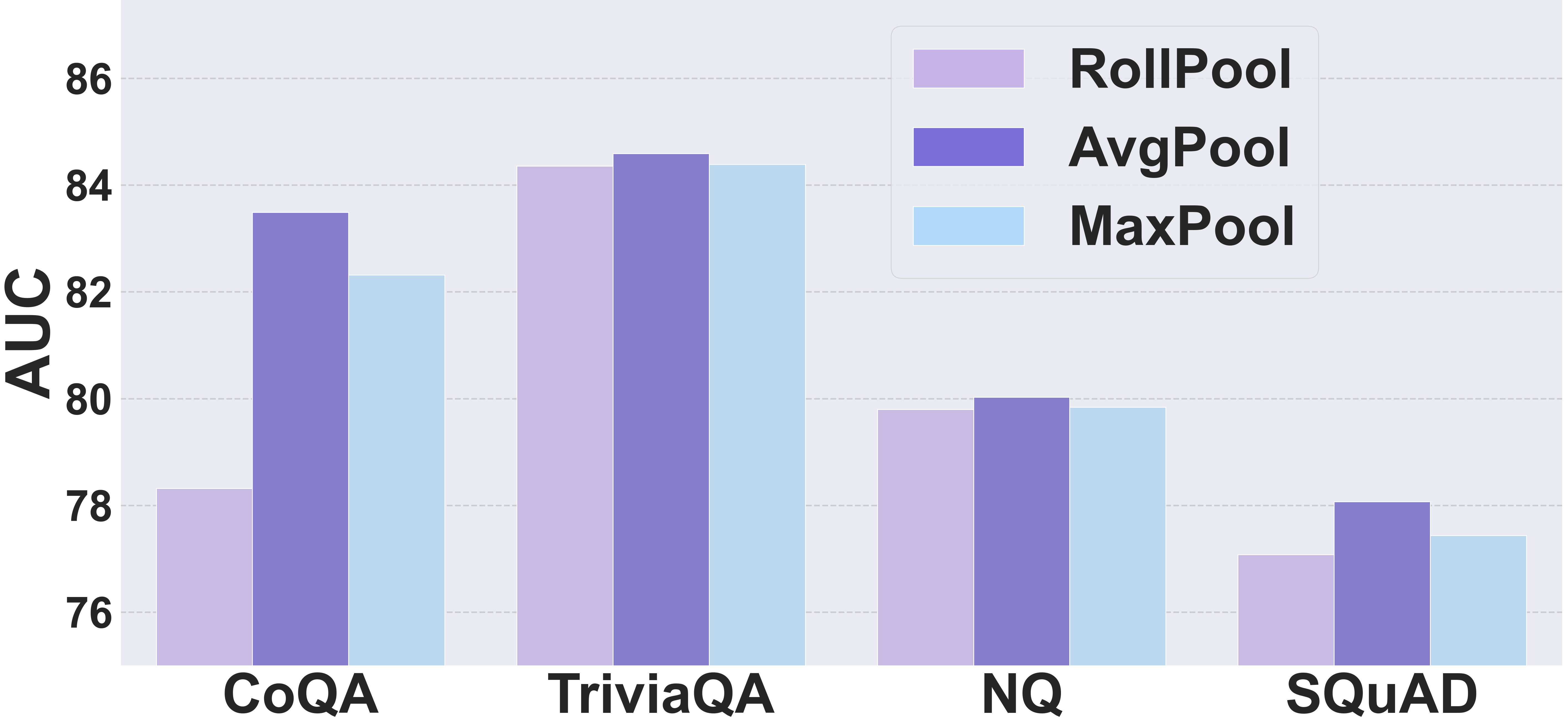}
    \caption{Comparison of base attribution strategies.}
    \label{fig:our_attr}
\end{figure}
\subsection{Ablation Studies}
\paragraph{Computation Variation.}
To further verify the effectiveness of our proposed method, we conduct ablation experiments on {Qwen2.5-7B-Instruct} and illustrate them in this section.
Firstly, we compare \textbf{InnerPPL} \& \textbf{OuterPPL} and the vanilla perplexity version of our method (multiplying original Perplexity in~\Cref{eq:7}) to prove the efficacy of our design.
As~\Cref{tab:ablation} demonstrated, \textbf{RePPL} certainly improves performance over all its computational variations.
Besides, \textbf{InnerPPL} and \textbf{RePPL} (Perplexity) derived from the original perplexity perform well individually.
\label{sec:4.3}
\begin{table}[!htbp]
\centering
\small
\setlength{\tabcolsep}{3.5pt}
\renewcommand{\arraystretch}{0.95}
\begin{tabular}{lcccc}
\toprule
\multirow{2}{*}{\textbf{Ablations}}
& \multicolumn{4}{c}{\textbf{Datasets}}  \\
\cmidrule(lr){2-5}
 & CoQA & TriviaQA & NQ & SQuAD \\
\midrule
InnerPPL & 83.2 & 80.1 & 76.3 & 75.3\\
OuterPPL & 78.2 & 84.1 & 77.7 & 76.1\\
RePPL {(Perplexity)} & 82.1 & 84.5 & 79.6 & 77.7 \\
\textbf{RePPL} & \textbf{83.5} & \textbf{84.6} & \textbf{80.0} & \textbf{78.1}\\
\bottomrule
\end{tabular}
\caption{Ablation studies on Qwen2.5-7B-Instruct.}
\label{tab:ablation}
\end{table}
\paragraph{Attribution Strategies.}
We also investigate the influence of different attribution methods in~\Cref{fig:our_attr}.
An unexpected finding is that \textbf{AvgPool} performs best across all comparisons.
Meanwhile, considering the lower computational complexity of \textbf{AvgPool}, we recommend it as the approach to derive the attribution map in our method.
\paragraph{Hyperparameters ($\alpha$, $\epsilon$).}
As for the hypermeter group $(\alpha, \epsilon)$, the ablation results are shown in~\Cref{fig:plot}(a).
It indicates that our method remains robust under different hyperparameter settings.
Given that $(1.0, \ 0.005)$ achieves the best overall performance across 4 datasets, we recommend it as the default setting of our method.
\subsection{Sensitivity to Sampling Hyperparameters}
In this section, we extend our analysis to sensitivity to randomness in sampling, which is controlled by the following decoding hyperparameters.
For efficiency considerations, a random one-fifth subset was taken from each dataset.
The observed performance differs by no more than 1.5\% from the main results.
As~\Cref{fig:plot} (b) (c) (d) shows, randomness at different sampling hyperparameters affects our performance.
However, this effect is limited as these curves display a moderately gradual ascent from lower randomness hyperparameters to higher ones. It indicates that our method is robust.

\section{Explainability Analysis}
\subsection{Faithfulness Test}
In line with previous XAI works~\citep{SelfAttr, GenAtt, AttnLRP}, we assess explanation faithfulness via perturbation tests that progressively mask low-scoring tokens. Here, we emphasize that explanation faithfulness is consistency with the model's nature, not human cognition.
The difference is that the assessment here focuses on which explanation more effectively highlights tokens that are more prone to inducing model hallucinations (uncertainty), rather than those that contribute more to the output (importance), hence, it needs detection performance as the indicator.
A smaller detection performance drop across masking ratios indicates greater faithfulness, and the absolute performance itself also reflects explanation reliability.
Here, we stress that this perturbation test is about vaildating the explanation faithfulness towards uncertainty-related hallucination nature, rather than fitness towards the human-reasonable cause of hallucination.
In other words, the explainability of our method is focusing on making LLM's uncertainty-based hallucination pattern readable and visualizible, instead of offering intuitively assumed explanation from human cognition.

\begin{table}[!htbp]
\centering
\small
\setlength{\tabcolsep}{3pt}
\renewcommand{\arraystretch}{0.9}
\begin{tabular}{lccccc}
\toprule
\multirow{2}{*}{\makecell[c]{\textbf{Explanation} \\ \textbf{Scores}}}
& \multicolumn{5}{c}{\textbf{TriviaQA}}  \\
\cmidrule(lr){2-6}
 & @0\% & @25\% & @50\% & @75\%  & Average\\
\midrule
Raw Logits & 66.9 & 67.1 & 65.5 & 63.7 & 65.8\\
AvgPool & 60.0 & 60.0 & 59.3 & 57.6 & 59.2\\
MaxPool & 62.6 & 62.7 & 62.3 & 62.4 & 62.5\\
RollPool & 55.4 & 55.3 & 55.2  & 55.0 & 55.2\\
\textbf{InnerPPL} & \textbf{79.7} & \textbf{79.4} & \textbf{79.1} & \textbf{78.6} & \textbf{79.2} \\
\midrule

\multirow{2}{*}{\makecell[c]{\textbf{} \\ \textbf{}}}
& \multicolumn{5}{c}{\textbf{SQuAD}}  \\
\cmidrule(lr){2-6}
 & @0\% & @25\% & @50\% & @75\%  &  Average\\
\midrule
Raw Logits & 51.2 & 51.0 & 50.8 & 50.2 & 50.8\\
AvgPool & 51.3 & 51.5 & 52.0 & 53.0 & 52.0\\
MaxPool  & 50.2 & 50.8 & 52.1 & 54.5 & 51.9\\
RollPool  & 48.0 & 48.2 & 48.9 & 50.5 & 49.0\\
\textbf{InnerPPL} & \textbf{74.2} & \textbf{73.9}& \textbf{73.8} & \textbf{73.2} & \textbf{73.8} \\
\bottomrule
\end{tabular}
\caption{Faithfulness (AUC) test on TriviaQA and SQuAD. 0\%\textasciitilde
75\% are different masking ratios.}
\label{tab:faith}
\end{table}

To assess the faithfulness of identifying uncertainty-inducing tokens, we conduct the test in the input space (corresponding to log rescaled uncertainty score that consists \textbf{InnerPPL}), using one-fifth random subsets from TriviaQA and SQuAD.
As for comparison, we take the raw logits in throughput of the input tokens, importance-oriented attribution scores of the three pooling strategies we adopt as explanation scores baselines.
We steply mask these hallucination scores at ratios of 0\%, 25\%, 50\%, and 75\% and use the AUC metric for evaluation.
~\Cref{tab:faith} depicts a slight decrease in our method's performance at different masking ratios, which proves the primary capability of rightfully attributing significant uncertainty-inducing tokens.
On the contrary, all baselines show erratic drops, which show their inability to explain.
Our method is even robust at the 75\% masking ratio. This suggests its capacity to truly concern uncertainty-inducing tokens of hallucination.
\begin{figure}[!htbp]
    \centering
    \includegraphics[width=0.95\linewidth]{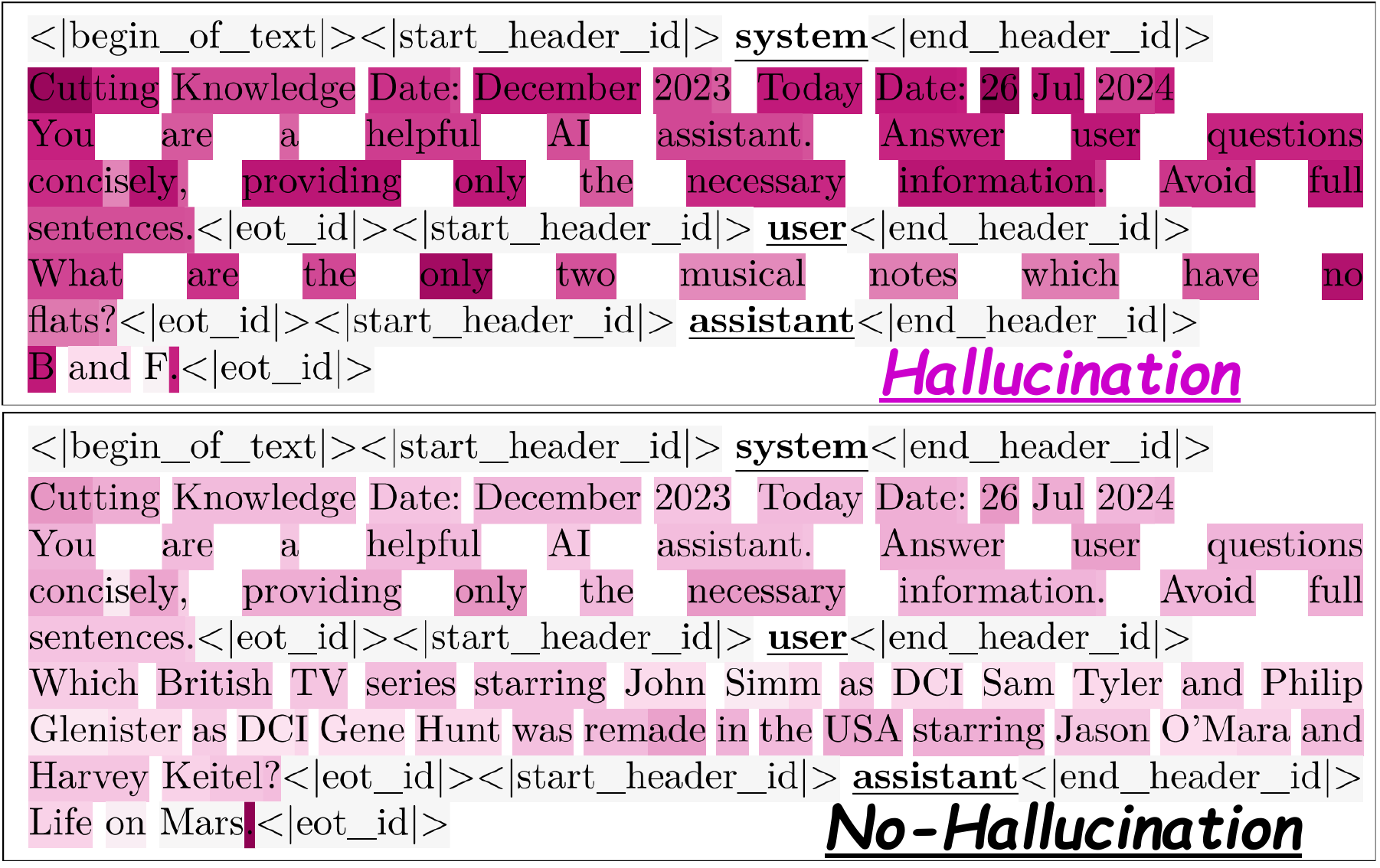}
    \caption{Visual explanations of the whole QA process on TriviaQA. The upper ones are detected as hallucinations by \textbf{RePPL}. The lower ones is non-hallucinated. Here, we mask the special token for better presentation.}
    \label{fig:visTQA}
\end{figure}
\subsection{Explanation for Hallucination}
\label{sec:5.2}
We concatenate uncertainty scores that consist \textbf{InnerPPL} and \textbf{OuterPPL} as a token-wise and visualizable hallucination explanation.
We take the log-rescaled uncertainty degree for visualization in~\Cref{fig:visTQA} and~\Cref{ap:mvis}, where color intensity represents a consistent range of values across examples. These visualizations demonstrate good distinctness, but do not fully accord with human cognition.
Besides these mirco-level visualization explanation, we explore the relation between attribution importance score and uncertainty explanation score at a marco-level in~\Cref{ap:relation}.

\section{Conclusion}
Motivated by the explainability limitations of the current hallucination detection method, we introduce a fresh perspective to exploit uncertainty in semantic propagation and language generation.
Based on that, we propose our \textbf{RePPL} to utilize the discrepancy between the attribution of different sampled generations and combine refined generation confidence.
These designs equip our method with our hierarchical distinctness for better detection and token-level uncertainty correspondence for explanation.
The experiments indicate that our method achieves strong detection performance, and it is able to highlight uncertainty-inducing tokens as an explanation.
We hope our insights, even drawbacks, can better motivate future research in this domain to better tackle the hallucination of LLM and build more trustworthy AI systems in future.

\clearpage
\section*{Limitations}
Our method is inherently limited by the following three drawbacks: 
(1) It mainly captures uncertainty-based hallucinations, whereas recent work~\citep{CanYouUncertain} highlights the existence of high-confidence hallucinations that slightly fall outside this range. 
(2) Its explanation is relatively coarse. This is partly due to the trade-off in time complexity, as the chosen attribution methods are different from general pooling of raw attention maps, which favors efficiency over the finer granularity offered by more computationally intensive state-of-the-art methods. 
(3) Our findings in~\Cref{ap:relation} are also preliminary, which mainly embodies its potential utility. It requires more comprehensive statistical investigations at a larger scale.

However, these drawbacks also point out valuable directions, such as unifying perturbation and sampling for more precise attribution. We believe our insights would gain better implementation by overcoming these limitations in future studies.
\bibliography{main}

\clearpage
\appendix

\section{Implementation Details}
\label{ap:impl}

We conduct all of our experiments on an NVIDIA H100 GPU, with full precision FP32. 

All of the implementations are based on \verb|transformers| Library of Huggingface\footnote{\url{https://github.com/huggingface/transformers}}. And all generations include related internal states and attention maps are derived from the \verb|.generate(...)| of \verb|GenerateMixin| Class (Which is the part of the Model Class). In addition, attention maps can be returned with setting with the \verb|output_attentions=True| and \verb| attn_implementation="eager"| when loading the model's weights. 

For model weights of Llama-3.1-8B-Instruct\footnote{\url{https://huggingface.co/meta-llama/Llama-3.1-8B-Instruct}}, Qwen2.5-7B-Instruct\footnote{\url{https://huggingface.co/Qwen/Qwen2.5-7B-Instruct}}, Qwen2.5-14B-Instruct\footnote{\url{https://huggingface.co/Qwen/Qwen2.5-14B-Instruct}}, and the sentence embedding model all-MiniLM-L6-v2\footnote{\url{https://huggingface.co/sentence-transformers/all-MiniLM-L6-v2}}, we choose the open-source weights from their Huggingface repositories. 

For the ROUGE metric, we select the standard implementation of Google Research\footnote{\url{https://github.com/google-research/google-research/tree/master/rouge}}.

\section{Inference Settings}
\label{ap:set}
Since we select the instruction-finetuned model for experiments, to more closely simulate practical scenarios, we adopt the chat template for generation. In the system prompt, we constrain the model to only answer key information by the instructions. As a result, the model's responses are closer to the standard answer of QA pairs. The input format of QA pairs is shown as follows:
\begin{tcolorbox}[
  enhanced,
  colback=black!3!white,
  colframe=black!40!white,
  fonttitle=\bfseries,
  title={Input Format (Without Context)},
  coltitle=black,
  colbacktitle=black!40!white,
  boxrule=0.8pt,
  top=4pt,
  bottom=4pt,
  left=5pt,
  right=5pt,
]
\textbf{System Prompt}: \textit{You are a helpful AI assistant. Answer user questions concisely, providing only the necessary information. Avoid full sentences. }\\ \textbf{User Prompt}: \texttt{<Question>}
\end{tcolorbox}
The input format with context is:
\begin{tcolorbox}[
  enhanced,
  colback=black!3!white,
  colframe=black!40!white,
  fonttitle=\bfseries,
  title={Input Format (Without Context)},
  coltitle=black,
  colbacktitle=black!40!white,
  boxrule=0.8pt,
  top=4pt,
  bottom=4pt,
  left=5pt,
  right=5pt,
]
\textbf{System Prompt}: \textit{You are a helpful AI assistant. Answer user questions based on provided context concisely, providing only the necessary information. Avoid full sentences.} \\ \textbf{User Prompt}: \textit{Context:} \texttt{<Context>} \textit{Question:} \texttt{<Question>}
\end{tcolorbox}

\section{Details of Metrics}
\label{ap:metric}
\paragraph{AUC.}  Area Under the Receiver Operating Characteristic Curve (AUC) is the classic metric for measuring the performance of a classifier, specifically the score-based classifier without a preset threshold. 
\paragraph{Acc.} Our accuracy (Acc) is the accuracy at the threshold spot that receives maximum \textbf{G-Mean}~\citep{Gmean} value. This is computed as $\textbf{G-Mean} = \sqrt{\text{TPR} \times (1- \text{FPR})}$, where $\text{TPR}$ is true positive rate and $\text{FPR}$ is false positive rate. When \textbf{G-Mean} reaches the maximum, the performance of the score-based classifier achieves its best at the corresponding threshold, especially in the imbalanced scenarios~\citep{Gmean}. In other words, the accuracy for predicting the positive class and the negative class has achieved the geometric mean optimal value. Hence, the accuracy at this specified threshold reflects the overall performance. 
\paragraph{Corr.} We select Spearman Correlation~\citep{Spearman} (Corr) as our performance metric. It non-linearly embodies the overall similarity between the hallucination detection score and the correctness measure score (soft version of hallucination ground truth) on the whole dataset.
\paragraph{PRR.} The Prediction Rejection Ratio~\citep{PRR} (PRR) is a metric designed to evaluate the effectiveness of an uncertainty quantification method by assessing how well uncertain predictions correspond to different quality responses of LLM (quality here is the score of the two correctness measures we adopt). It firstly computes from a prediction-rejection curve, which plots the average quality score of the correctness measure when the most uncertain predictions (the biggest uncertain scores for hallucination detection) are progressively discarded.
Let $\mathrm{Area}_{\text{unc}}$ denote the area under this rejection curve induced by the uncertainty scores.
In contrast, we also plot the curve that first discards the lowest quality score as an optimal rejection strategy curve for comparison. Let $\mathrm{Area}_{\text{oracle}}$ denote the area under the optimal rejection strategy that discards the lowest-quality samples first. PRR is defined as ($\mathrm{Area}_{\text{rnd}}$ is the area for a randomly permuted uncertainty score):
\begin{align}
\mathrm{PRR} = \frac{\mathrm{Area}_{\text{unc}} - \mathrm{Area}_{\text{rnd}}}{\mathrm{Area}_{\text{oracle}} - \mathrm{Area}_{\text{rnd}}}. \notag
\end{align}

\section{Baseline Implemention}
\label{ap:baseline}
\paragraph{Perplexity.} We compute perplexity~\citep{Perplexity} as the following equation:
\begin{equation}
    \textbf{PPL} = - \frac{1}{S} \sum_i^S log(p), \notag
\end{equation}
here, $S$ is the length of generation and $p$ is logits of generation tokens.
\paragraph{Energy.} Energy score~\citep{EnergyScore} is the result of exp-log rescaling:
\begin{equation}
    \textbf{Energy} = \frac{1}{S} \sum_{i=1}^{S} log( \sum_{j} \exp (p)), \notag
\end{equation}
which protrude the most uncertain logits.
\paragraph{P(True).} It gives LLM a prompt to let LLM answer the question about whether the candidate answers hallucinate~\citep{P(True)}. We prompt LLM as follows:
\begin{tcolorbox}[
  enhanced,
  colback=blue!3!white,
  colframe=blue!40!white,
  fonttitle=\bfseries,
  title={Input Format (Without Context)},
  coltitle=black,
  colbacktitle=blue!40!white,
  boxrule=0.8pt,
  top=4pt,
  bottom=4pt,
  left=5pt,
  right=5pt,
]
\textbf{System Prompt}: \textit{You are a helpful assistant. You are asked to determine whether the possible answer correctly responds to the entire question, which may include context.}\\ \textbf{User Prompt}: \textit{Entire Question (may include context):} <Question> \\ \textit{Possible answer:} <Candidate Response>\\ Does the possible answer correctly respond to the question above? Answer `Yes' or `No' only:
\end{tcolorbox}
Then, if the "Yes" token (or equivalent tokens) appears in answers, we count the probability of this token $ P(y=\text{``Yes"})$ as the hallucination score for detection.
\paragraph{Length-Normalized Predictive Entropy (LNPE).} LNPE~\cite{LNPE} proposes to measure sequence-level uncertainty by calculating the expectation of uncertainty in multiple generations, which is mathematically equivalent to the average of the PPL across different generations. It is computed as:
\begin{equation}
    \textbf{LNPE} = - \frac{1}{N} \sum_{m=1}^{N} \frac{1}{S_n} \sum_{t=1}^{S_n} \log (p) \notag.
\end{equation}
Here, $1, 2, ...,n$ is the index of each generation in N times and the $S_n$ is the corresponding generation length.
\paragraph{Semantic Entropy.} The core idea of this method is to cluster the uncertainty of similar outputs then accumulate overall uncertainty that avoids distortion in the direct average accumulation~\cite{SemanticEnt}. It is:
\begin{align}
    & H_m = - \sum_{i=1}^{C_m} p_i \log p_i \ , \notag \\
    & \textbf{Semantic Entropy} = \frac{1}{M} \sum_{j=1}^{m} H_j \ . \notag
\end{align}
Therein, we have $M$ clusters $1, 2, ..., m$, each has $C_m$ numbers of outputs which contain $H_m$ semantic entropy of the cluster.
\paragraph{EigenScore.}
This estimates the hallucination degree of generated outputs based on the semantic dispersion in the sentence embedding space~\citep{EigenScore}. It computes the covariance matrix of $N$ generated sentence embeddings and then uses the log and determinant of the regularized covariance matrix to capture the diversity and confidence of the generation. It is defined as:
\begin{equation}
    \textbf{EigenScore} = \frac{1}{N} \log \det(\Sigma + \alpha \cdot \mathbf{I}_N) \notag,
\end{equation}
where $\Sigma = \mathbf{Z}^\top \mathbf{J}_d \mathbf{Z} \in \mathbb{R}^{N \times N}$ is the centered covariance matrix of $N$ sentence embeddings $\mathbf{Z} = [z_1, \dots, z_n] \in \mathbb{R}^{d \times N}$, $\mathbf{J}_d = \mathbf{I}_d - \frac{1}{N} \mathbf{I_N}\mathbf{I_N}^T$, and $\alpha \cdot \mathbf{I}_N$ is a small regularization term added to make the covariance matrix full rank. $z_n$ is the middle layer hidden state of the last token.
\paragraph{AttnScore.}
AttnScore is a method that leverages internal components of the self-attention mechanism to detect hallucinations, which is the best method in the LLM-Check toolbox~\citep{LLM_Check}. Instead of using hidden states, it analyzes the lower-triangular kernel matrices from each self-attention head. 
For a given input-output sequence $x_p \oplus x$ of length $m$, let $\text{Ker}^i \in \mathbb{R}^{m \times m}$ denote the lower-triangular kernel matrix of the self-attention map from attention head $i \in \{1, \dots, a\}$. Each matrix is computed via a Softmax over scaled dot products and has non-negative diagonal entries. Since these diagonals match the eigenvalues of $\text{Ker}^i$, we compute the log-determinant of each head as the sum of logarithms of its diagonal entries. The final AttnScore is obtained by averaging these log-determinants across all $a$ heads and normalizing by the sequence length $m$:
\begin{equation}
\textbf{AttnScore} = \frac{1}{a \cdot m} \sum_{i=1}^{a} \sum_{j=1}^{m} \log \text{Ker}^i_{jj}. \notag
\end{equation}

\paragraph{SAPLMA.}
SAPLMA~\cite{SAPLMA} is a method for hallucination prediction that leverages the last hidden state (corresponding to the last token) activations at middle-high layers from LLMs. Specifically, we train a feedforward neural network classifier with a 3-layer MLP architecture (dimensions decrease exponentially)  for 10 epochs using the Adam optimizer and the Binary Cross Entropy Loss, with the AUC metric being the proof of saving the optimal checkpoint. All data is collected from the train set of the dataset we adopt (we illustrate in the main text that we use the validation set of this dataset for performance comparison), at a size of 4 times the validation set we use for comparison (all parameterized methods use this same training data).

\paragraph{Probe-LR.}
It uses a logistic regression (LR) probe~\citep{ProbeLR} at the last hidden state (corresponding to the last token) activations at middle layers from LLMs to detect hallucinations. Given a prompt and generated response, we extract the hidden states in the middle layer from the LLM during generation. We adopt the same training strategy of SAPLMA to train it (the optimal learning rate is different).

\paragraph{Probe-MM.}
To distinguish between hallucinated (H) and non-hallucinated (NH) instances in the training set, we first derive the average activations for each group separately. That is the core idea of the Mass-Mean-style (MM) probe~\citep{ProbeMM}. Specifically, we take the last token in the middle layer, we compute the average response $\mu$ over all hallucinated training samples $\mathcal{D}_{\text{train}}^{\text{H}}$, and $\nu$ over all non-hallucinated samples $\mathcal{D}_{\text{train}}^{\text{NH}}$:
\begin{equation}
\begin{aligned}
\mu &= \frac{1}{|\mathcal{D}_{\text{train}}^{\text{H}}|}
\sum_{t \in \mathcal{D}_{\text{train}}^{\text{H}}} x^H_t, \\
\nu &= \frac{1}{|\mathcal{D}_{\text{train}}^{\text{NH}}|}
\sum_{q \in \mathcal{D}_{\text{train}}^{\text{NH}}} x^{NH}_q. \notag
\end{aligned}
\end{equation}
Next, we define a contrastive signal by taking the difference between the mean activations of the H and NH classes:

\begin{equation}
\theta = \mu - \nu. \notag
\end{equation}
Finally, we can use $\theta$ as the weights of the linear probe $p_{halu}(x) = \theta x$ to judge whether the $x$ corresponding response is hallucinated or not.

\begin{table*}[!t]
\resizebox{1\textwidth}{!}
{
\centering
\small
\setlength{\tabcolsep}{0.9pt}
\renewcommand{\arraystretch}{1.1}
\begin{tabular}{llcccccccccccccccccccc}
\toprule
\multirow{2}{*}{\textbf{Models}} & \textbf{Datasets}
& \multicolumn{4}{c}{\textbf{CoQA}}
& \multicolumn{4}{c}{\textbf{TriviaQA}}
& \multicolumn{4}{c}{\textbf{NQ}}
& \multicolumn{4}{c}{\textbf{SQuAD}}
& \multicolumn{4}{c}{\textbf{ALL}} \\
\cmidrule(lr){3-6} \cmidrule(lr){7-10} \cmidrule(lr){11-14} \cmidrule(lr){15-18} \cmidrule(lr){19-22}
& \textbf{Methods} & AUC & Acc & Corr & PRR
& AUC & Acc & Corr & PRR
& AUC & Acc & Corr & PRR
& AUC & Acc & Corr & PRR 
& AUC & Acc & Corr & PRR \\
\midrule
\multirow{11}{*}{\makecell[c]{\textbf{LLaMA-3.1-} \\ \textbf{8B-Instruct}}}
& Perplexity & 70.4 & 65.0 & 34.6 & 44.6 & 78.5 & 71.3 & 53.7 & 62.9 & 74.8 & 69.1 & 42.9 & 49.3 & 71.2 & 65.8 & 33.8 & 30.6 & 76.2 & 69.5 & 47.1 & 53.3 \\
& Energy & 68.1 & 62.1 & 30.9 & 41.7 & 68.0 & 63.6 & 35.6 & 40.5 & 65.3 & 62.7 & 28.8 & 32.2 & 70.7 & 66.2 & 35.4 & 47.2 & 71.9 & 66.0 & 39.7 & 48.8 \\
& P(True) & 66.4 & 61.6 & 29.6 & 39.3 & 77.9 & 71.1 & 50.7 & 60.0 & 69.3 & 65.8 & 35.6 & 38.9 & 68.6 & 61.0 & 27.5 & 38.5 & 74.4 & 68.1 & 44.0 & 52.6 \\
& LNPE & 75.4 & 68.1 & 43.6 & 51.9 & 81.0 & 73.4 & 56.9 & 66.5 & \textbf{76.1}& \textbf{69.5} & \textbf{43.1} & \textbf{50.0} & 73.5 & 70.7 & 37.7 & 35.5 & 79.3 & 72.4 & 52.1 & 57.4 \\
& Semantic Entropy & 75.3 & 70.0 & 44.2 & 52.2 & 75.2 & 69.2 & 44.3 & 53.4 & 66.1 & 64.1 & 25.7 & 31.4 & 71.6 & 73.2 & 33.3 & 35.8 & 76.1 & 70.7 & 45.5 & 52.7 \\
& EigenScore & 79.0 & 73.8 & 49.0 & 58.6 & 80.8 & 73.7 & 55.2 & 65.1 & 71.9 & 70.4 & 33.6 & 42.4 & \textbf{80.4} & \textbf{73.4} & \textbf{51.9} & \textbf{62.7} & 79.2 & 71.2 & 51.1 & 60.6 \\
& AttnScore & 71.9 & 68.5 & 38.7 & 37.9 & 72.9 & 67.0 & 40.9 & 44.5 & 62.5 & 59.8 & 16.4 & 24.8 & 67.2 & 64.5 & 28.8 & 28.8 & 59.7 & 54.6 & 17.1 & 25.3 \\
& \textbf{RePPL (Ours)} & \textbf{83.5} & \textbf{77.0} & \textbf{60.6} & \textbf{67.0} & \textbf{85.2} & \textbf{76.9} & \textbf{62.5} & \textbf{71.1} & 75.3 & 69.4 & 37.9 & 48.6 & 76.0 & 70.8 & 48.1 & 40.6 & \textbf{82.8} & \textbf{75.5} & \textbf{58.9} & \textbf{63.6} \\
& SAPLMA* & \underline{88.5} & \underline{80.7} & \underline{69.2} & 78.4 & \underline{90.2} & \underline{82.5} & 68.1 & \underline{78.4} & 72.9 & 67.2 & 34.9 & 45.1 & \underline{87.2} & \underline{79.3} & \underline{59.0} & \underline{72.6} & 88.2 & 80.5 & 66.0 & 75.2 \\
& Probe-LR* & 88.1 & 80.1 & 69.2 & \underline{79.0} & 90.0 & 82.3 & \underline{68.7} & 78.3 & \underline{76.5} & \underline{69.4} & \underline{41.1} & \underline{51.8} & 86.8 & 79.0 & 58.3 & 69.6 & \underline{88.6} & \underline{80.6} & \underline{67.8} & \underline{76.7} \\
& Probe-MM* & 79.6 & 72.4 & 55.4 & 61.9 & 83.4 & 75.3 & 60.6 & 71.1 & 70.1 & 63.9 & 29.6 & 37.4 & 81.6 & 76.4 & 50.2 & 62.3 & 73.9 & 67.9 & 42.8 & 46.4 \\
\midrule
\multirow{11}{*}{\makecell[c]{\textbf{Qwen-2.5-} \\ \textbf{7B-Instruct}}}
& Perplexity & 74.4 & 67.6 & 44.9 & 56.2 & 83.5 & 75.6 & 60.4 & 70.4 & 75.5 & 69.6 & \textbf{38.8} & 41.7 & 74.1 & 69.1 & 46.9 & 49.6 & 81.2 & 73.0 & 58.1 & 65.6 \\
& Energy & 60.2 & 57.3 & 16.1 & 21.7 & 77.6 & 70.8 & 49.4 & 52.5 & 64.6 & 57.5 & 27.6 & 22.3 & 60.8 & 58.7 & 15.9 & 19.4 & 71.0 & 65.8 & 39.8 & 42.1 \\
& P(True) & 57.7 & 57.2 & 16.4 & 15.8 & 80.8 & 75.0 & 57.3 & 65.4 & 65.1 & 63.9 & 26.5 & 32.3 & 59.2 & 60.3 & 14.2 & 20.4 & 74.1 & 69.8 & 46.4 & 49.6 \\
& LNPE & 76.3 & 69.9 & 48.0 & 58.5 & 83.7 & 76.0 & 60.6 & \textbf{70.8} & 76.4 & 66.6 & 40.1 & \textbf{44.3} & 75.1 & 70.2 & 48.1 & 50.5 & 82.1 & 74.3 & 59.5 & 66.1 \\
& Semantic Entropy & 74.2 & 73.8 & 40.9 & 40.2 & 77.7 & 76.0 & 49.4 & 54.7 & 71.8 & \textbf{73.1} & 30.3 & 36.4 & 65.9 & \textbf{79.3} & 22.2 & 24.3 & 77.7 & 76.4 & 49.0 & 50.9 \\
& EigenScore & \textbf{81.7} & \textbf{76.0}& 56.4 & 58.4 & 82.6 & 75.9 & 57.7 & 60.8 & 77.4 & 72.1 & 37.4 & 41.0 & 78.3 & 74.6 & 50.6 & \textbf{53.3} & 82.8 & \textbf{76.5} & 58.3 & 62.2 \\
& AttnScore & 71.0 & 65.9 & 34.6 & 39.6 & 60.9 & 58.0 & 18.7 & 18.8 & 62.2 & 56.2 & 13.8 & 14.1 & 71.1 & 65.3 & 35.8 & 42.2 & 60.8 & 59.4 & 21.8 & 31.0 \\
& \textbf{RePPL (Ours)} & 81.2 & 73.7 & \textbf{57.1} & \textbf{64.0} & \textbf{84.1} & \textbf{76.0} & \textbf{61.1} & 70.2 & \textbf{77.7} & 67.0 & 37.8 & 43.8 & \textbf{77.1} & 70.5 & \textbf{51.6 }& 52.6 & \textbf{83.3} & 75.4 & \textbf{60.8} & \textbf{67.7} \\
& SAPLMA* & \underline{91.2} & \underline{83.0} & \underline{74.3} & \underline{82.6} & \underline{90.6} & \underline{83.1} & \underline{71.7} & \underline{81.8} & \underline{83.4} & \underline{73.6} & \underline{45.3} & \underline{57.6} & \underline{88.0} & \underline{78.3} & \underline{65.8} & \underline{74.6} & \underline{90.4} & \underline{83.0} & \underline{70.6} & \underline{79.2} \\
& Probe-LR* & 88.3 & 80.4 & 70.0 & 76.6 & 86.8 & 78.8 & 65.3 & 74.7 & 79.7 & 72.6 & 40.1 & 49.2 & 83.4 & 76.0 & 59.1 & 66.2 & 87.8 & 79.9 & 67.2 & 75.6 \\
& Probe-MM* & 73.9 & 68.9 & 43.8 & 48.9 & 76.6 & 71.5 & 50.2 & 50.4 & 74.8 & 71.0 & 35.0 & 34.3 & 76.3 & 71.5 & 49.2 & 49.8 & 73.4 & 67.8 & 41.2 & 39.8 \\
\midrule
\multirow{11}{*}{\makecell[c]{\textbf{Qwen-2.5-} \\ \textbf{14B-Instruct}}}
& Perplexity & 72.8 & 66.9 & 44.9 & 48.2 & 80.8 & 73.2 & 57.0 & 66.0 & 74.3 & 68.0 & 38.7 & 44.1 & 75.0 & 67.7 & 49.2 & 51.8 & 77.0 & 70.3 & 52.6 & 57.8 \\
& Energy & 61.1 & 57.0 & 27.4 & 19.1 & 63.5 & 61.0 & 28.4 & 26.4 & 60.0 & 56.7 & 22.8 & 21.9 & 60.0 & 58.8 & 20.8 & 21.4 & 57.4 & 54.1 & 22.8 & 19.4 \\
& P(True) & 60.8 & 57.9 & 27.0 & 21.4 & 72.4 & 71.0 & 44.8 & 40.9 & 62.0 & 65.2 & 28.3 & 25.6 & 54.0 & 55.9 & 9.4 & 2.4 & 63.1 & 62.6 & 33.8 & 23.4 \\
& LNPE & 74.2 & 68.0 & 47.4 & 49.2 & 81.9 & 74.6 & 58.7 & 66.1 & 75.2 & 68.9 & \textbf{39.7} & 43.2 & 76.4 & 69.8 & 51.4 & 53.5 & 78.1 & 71.2 & 54.6 & 58.8 \\
& Semantic Entropy & 67.8 & 62.5 & 36.1 & 23.7 & 76.7 & 76.3 & 47.0 & 47.0 & 69.9 & 72.2 & 30.3 & 33.5 & 66.4 & 72.3 & 29.0 & 24.7 & 71.8 & 70.9 & 41.8 & 36.5 \\
& EigenScore & \textbf{80.1} & \textbf{72.2} & \textbf{56.0} & 51.1 & 82.6 & \textbf{77.2} & 57.1 & 60.7 & 76.4 & 72.2 & 37.0 & 37.3 & \textbf{81.5} & \textbf{74.4} & 58.5 & 57.4 & \textbf{81.4} & \textbf{74.9} & 56.0 & 56.3 \\
& AttnScore & 53.2 & 51.6 & 9.8 & 6.4 & 63.2 & 60.1 & 24.6 & 26.5 & 63.0 & 59.8 & 20.5 & 19.0 & 55.1 & 53.5 & 9.0 & 3.6 & 60.7 & 59.3 & 12.9 & 23.9 \\
& \textbf{RePPL (Ours)} & 76.3 & 71.2 & 50.6 & \textbf{52.7} & \textbf{84.1}& 76.2 & \textbf{61.1} & \textbf{68.6} & \textbf{77.5} & \textbf{69.2} & 39.1 & \textbf{44.3} & 80.6 & 72.6 & \textbf{58.9} & \textbf{58.0} & 80.7 & 73.4 & \textbf{56.9} & \textbf{61.2} \\
& SAPLMA* & \underline{90.8} & \underline{82.7} & \underline{71.9} & \underline{82.5} & \underline{92.6} & \underline{85.1} & \underline{73.7} & \underline{83.2} & \underline{81.9} & \underline{73.0} & \underline{43.5} & \underline{55.7} & \underline{91.9} & \underline{86.3} & \underline{77.4} & \underline{83.1} & \underline{91.4} & \underline{83.8} & \underline{72.8} & \underline{81.7}
 \\
& Probe-LR* & 89.4 & 82.4 & 69.1 & 80.1 & 90.2 & 82.7 & 69.8 & 78.9 & 78.9 & 72.4 & 38.9 & 48.3 & 91.1 & 85.3 & 73.7 & 78.2 & 89.7 & 82.1 & 68.1 & 77.7 \\
& Probe-MM* & 74.1 & 72.9 & 43.4 & 45.0 & 55.9 & 61.8 & 22.3 & 13.0 & 54.7 & 35.5 & 12.8 & 3.5 & 68.3 & 77.4 & 47.6 & 26.1 & 64.1 & 62.8 & 23.8 & 12.7 \\
\bottomrule
\end{tabular}
}
\caption{Evaluation results of different methods on various datasets using AUC, accuracy at threshold of max G-Mean (Acc), and Spearman Correlation (Corr) metrics. ROUGE-L (f1-measure) is the correctness measure. All values are percentages. The best performances of non-parameterized methods are in bold, and the best performances of parameterized methods (*) are underlined.}
\label{tab:main_results_rouge}
\end{table*}
\section{ROUGE-based Comparison}
\label{ap:rouge}
The results of ROUGE-L (f-measure) as the correctness measure are listed in~\Cref{tab:main_results_rouge}, with the threshold 0.5. Our method achieves the best comprehensive performance over individual datasets and the merge of 4 datasets with different models.

\begin{table*}[!htbp]

\resizebox{1\textwidth}{!}
{
\centering
\small
\setlength{\tabcolsep}{1.4pt}
\renewcommand{\arraystretch}{1.2}
\begin{tabular}{llcccccccccccccccccccc}
\toprule
\multirow{2}{*}{\textbf{Models}} & \textbf{Datasets}
& \multicolumn{4}{c}{\textbf{CoQA}}
& \multicolumn{4}{c}{\textbf{TriviaQA}}
& \multicolumn{4}{c}{\textbf{NQ}}
& \multicolumn{4}{c}{\textbf{SQuAD}}
& \multicolumn{4}{c}{\textbf{ALL}} \\
\cmidrule(lr){3-6} \cmidrule(lr){7-10} \cmidrule(lr){11-14} \cmidrule(lr){15-18} \cmidrule(lr){19-22}
& \textbf{Methods} & AUC & Acc & Corr & PRR
& AUC & Acc & Corr & PRR
& AUC & Acc & Corr & PRR
& AUC & Acc & Corr & PRR 
& AUC & Acc & Corr & PRR \\
\midrule
\multirow{4}{*}{\makecell[c]{\textbf{LLaMA-3.1-} \\ \textbf{8B-Instruct}}}
& \textbf{RePPL (Ours)} & 85.2 & 77.4 & 61.8 & 70.7 & 86.0 & 78.6 & 63.2 & 73.9 & 79.3 & 69.9 & 47.4 & 51.9 & 76.0 & 71.7 & 45.1 & 40.6 & 83.6 & 76.3 & 60.0 & 66.4 \\
& SAPLMA* & 88.5 & 81.1 & 66.7 & 80.1 & 87.7 & 79.3 & 67.6 & 79.4 & 74.3 & 69.6 & 40.4 & 45.6 & 79.8 & 73.2 & 52.5 & 70.7 & 85.6 & 78.4 & 64.4 & 75.4 \\
& Probe-LR* & 89.4 & 82.2 & 67.9 & 81.7 & 88.5 & 80.5 & 68.9 & 80.7 & 79.1 & 70.3 & 45.9 & 52.1 & 79.0 & 72.4 & 51.7 & 65.6 & 87.1 & 79.9 & 66.7 & 77.7 \\
& Probe-MM* & 85.0 & 77.7 & 62.1 & 73.8 & 86.5 & 78.1 & 63.4 & 75.4 & 77.1 & 67.6 & 31.2 & 36.2 & 75.3 & 68.3 & 46.2 & 61.8 & 77.5 & 70.4 & 48.8 & 53.4 \\
\midrule
\multirow{4}{*}{\makecell[c]{\textbf{Qwen-2.5-} \\ \textbf{7B-Instruct}}}
& \textbf{RePPL (Ours)} & 83.5 & 77.0 & 58.4 & 67.2 & 84.6 & 76.3 & 62.4 & 73.3 & 80.0 & 73.6 & 45.8 & 46.5 & 78.1 & 71.9 & 47.8 & 54.9 & 83.7 & 75.9 & 60.9 & 68.9 \\
& SAPLMA* & 92.0 & 84.7 & 73.1 & 82.3 & 89.8 & 82.5 & 71.7 & 81.9 & 84.7 & 75.5 & 50.5 & 56.8 & 84.1 & 78.2 & 60.0 & 72.6 & 89.4 & 82.5 & 70.8 & 78.3 \\
& Probe-LR* & 90.9 & 83.5 & 70.2 & 77.6 & 86.0 & 78.0 & 64.8 & 74.9 & 81.3 & 73.5 & 42.9 & 48.9 & 80.5 & 75.1 & 53.6 & 63.7 & 87.7 & 80.0 & 66.9 & 75.2 \\
& Probe-MM* & 80.5 & 73.6 & 52.3 & 57.5 & 77.2 & 72.6 & 53.5 & 55.6 & 76.9 & 69.9 & 30.3 & 30.8 & 73.9 & 71.0 & 46.2 & 52.9 & 75.9 & 70.6 & 47.4 & 45.9 \\
\midrule
\multirow{4}{*}{\makecell[c]{\textbf{Qwen-2.5-} \\ \textbf{14B-Instruct}}}
& \textbf{RePPL (Ours)} & 79.7 & 76.6 & 43.0 & 47.9 & 84.8 & 76.8 & 62.2 & 71.9 & 80.0 & 71.7 & 51.4 & 50.1 & 81.3 & 74.4 & 54.7 & 56.8 & 82.5 & 75.7 & 56.0 & 60.4 \\
& SAPLMA* & 96.5 & 92.0 & 68.5 & 79.0 & 90.7 & 82.9 & 73.4 & 83.4 & 83.7 & 77.1 & 54.9 & 58.3 & 90.2 & 83.4 & 72.3 & 78.8 & 91.5 & 84.3 & 73.1 & 80.8 \\
& Probe-LR* & 95.4 & 88.6 & 66.9 & 77.2 & 88.8 & 80.9 & 70.2 & 80.3 & 80.5 & 71.9 & 49.4 & 50.8 & 87.8 & 80.9 & 68.6 & 74.2 & 90.6 & 82.4 & 71.4 & 78.4 \\
& Probe-MM* & 85.0 & 78.9 & 43.9 & 42.0 & 55.2 & 50.5 & 26.7 & 17.2 & 54.6 & 25.5 & 24.3 & 6.8 & 64.5 & 62.0 & 40.9 & 22.1 & 66.3 & 56.2 & 38.0 & 20.7 \\
\bottomrule
\end{tabular}
}
\caption{Extra evaluation results of comparison between different parameterized methods (*) and ours on various datasets, using AUC, accuracy at threshold of max G-Mean (Acc), and Spearman Correlation (Corr) metrics. Embedding similarity is the correctness measure. All values are in percentages.}
\label{tab:aux_results_sen_emb}
\end{table*}
\section{Comparsion with Parameterized Probe-based Method}
\label{ap:probe}
We present comparison results with parameterized probe methods in~\Cref{tab:main_results_rouge} and~\Cref{tab:aux_results_sen_emb}. Our non-parametrized method is comparable to these strong methods. Given that these methods need extra engineering and are unable to work as uncertainty (hallucination risk) comparators without training, our non-parameterized method gains a good trade-off to be an effective uncertainty indicator with rich explainability.

\begin{figure*}[!t]
    \includegraphics[width=1.0\linewidth]{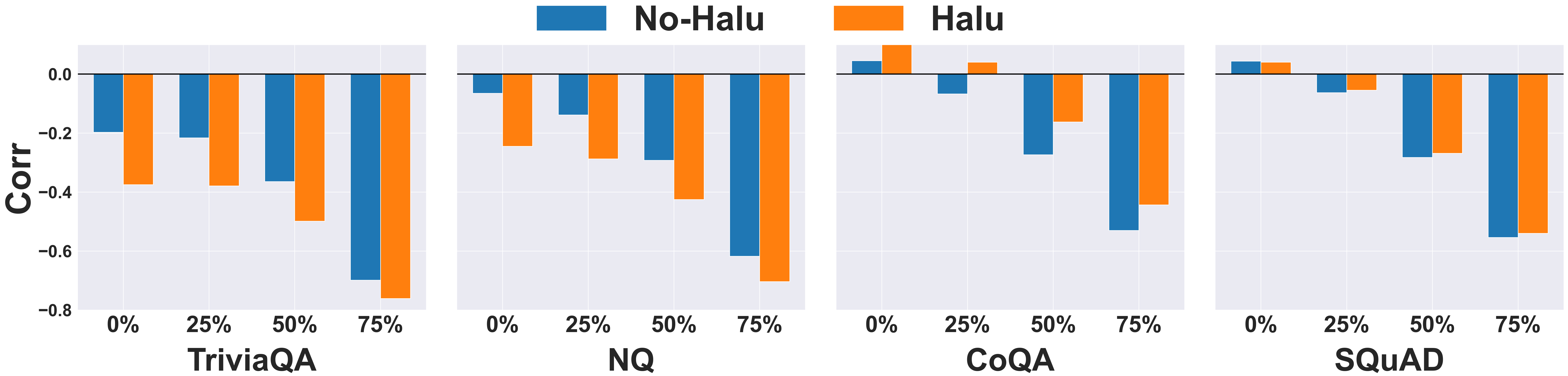}
    \caption{Average Spearman Correlation (Corr) between importance and uncertainty. X-axis correspond to different masking ratios. As can be seen, the negative average correlation is large at 75\% masking ratios. On the QA datasets with context, this negative correlation is smaller in hallucinated ones while larger on the datasets without context.}
    \label{fig:box}
\end{figure*}

\section{Runtime Analysis}
\label{ap:runtime}
We conduct a supplementary runtime analysis across four sampling-based methods: LNPE, Semantic Entropy, EigenScore, and our approach, on 100 randomly selected examples from TriviaQA and SQuAD using LLaMA-3.1-8B-Instruct. Each example involved 10 sampled outputs generated by the LLM. All methods are accounted for necessary forward passes, and inference is batched on 2 NVIDIA RTX 4090 GPUs (48GB total memory). The runtime results are summarized in~\Cref{tab:runtime}.

\begin{table}[!htbp]
\centering
\begin{tabular}{lcc}
\toprule
\textbf{Method} & \textbf{TriviaQA} & \textbf{SQuAD} \\
\midrule
LNPE & 65.58s & 205.27s \\
Semantic Entropy & 110.06s & 232.38s \\
EigenScore & 65.28s & 204.92s \\
\textbf{RePPL (Ours)} & 74.28s & 260.42s \\
\bottomrule
\end{tabular}
\caption{Runtime (in seconds, s) on 100 examples.} \label{tab:runtime}
\end{table}

As shown, our method maintains a runtime of approximately 0.65s and 2.6s per example on TriviaQA and SQuAD, which is reasonable among sampling-based approaches. While a bit slower than these three baselines, we argue the added explainability justifies the cost. In other words, we regard this trade-off as worthy since we bring better performance and explainability for hallucination detection.

\section{More Visual Explanations}
\label{ap:mvis}
\begin{figure}[!t]
    \centering
    \includegraphics[width=1.0\linewidth]{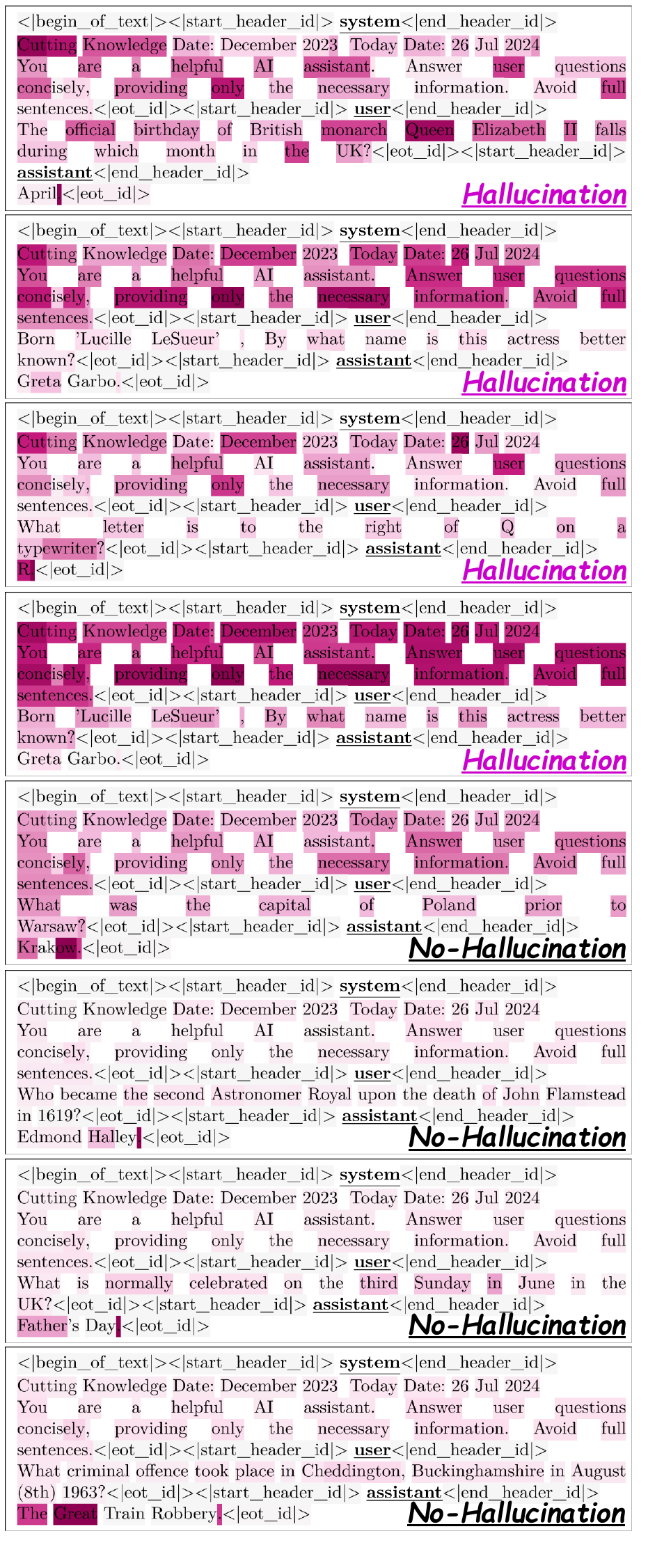}
    \caption{Visual explanations of the whole QA process on TriviaQA. The upper 4 are hallucinated ones, while the lower 4 are non-hallucinated ones.}
    \label{fig:whole_tqa}
\end{figure}

\begin{figure}[!t]
    \centering
    \includegraphics[width=1.0\linewidth]{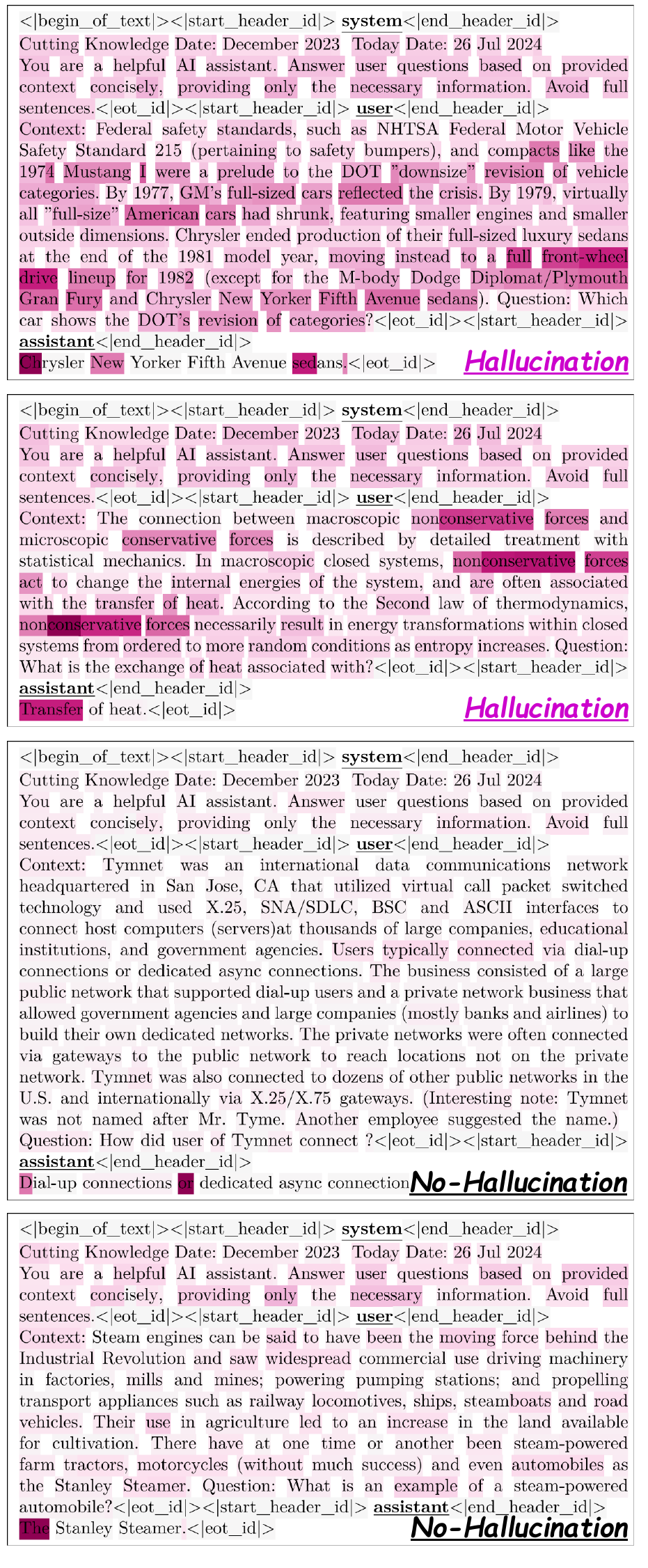}
    \caption{Visual explanations of the whole QA process on SQuAD. The Upper 2 are hallucinated ones, while the lower 2 are non-hallucinated ones.}
    \label{fig:whole_squad}
\end{figure}

We display 8 visual explanations on TriviaQA in~\Cref{fig:whole_tqa} and 4 visual explanations on SQuAD in~\Cref{fig:whole_squad} (we mask the specific tokens for better presentation).
As these figures show, distinctness between the hallucinated answers and the non-hallucinated answers is obvious. It also indicates that it is hard to detect hallucination with uncertainty in language generation alone.

\section{Importance and Uncertainty}
\label{ap:relation}
Coincidentally, during the computation process of our method, we first gain attribution-based importance and then derive the uncertainty about hallucination.
We computed the average Spearman Correlation between importance scores and uncertainty scores on the same data of ablation studies. or Spearman Correlation computing, the significant value $p > 0.05$ ones are treated as 0 correlation, and when both the uncertainty score and importance score are masked as 0 at the same time for the same token, this pair is removed for correct computation.
Results in~\Cref{fig:box} demonstrate two noteworthy findings:
(1) \textbf{Uncertainty is negatively correlated to importance.}
The negative correlation indicates uncertainty that induces hallucination, bringing chaos to the correct attention pattern.
(2) \textbf{This correlation corresponds to different hallucination types.} The different average correlation on the dataset with/without context illustrates a discrepant hallucination pattern about factual (external) and contextual (internal) ones.
Overall, these findings describe the disorder in the chaotic attention pattern when LLM tends to hallucinate. And demonstrate the potential usage of our explanation scores as a viable tool.
\end{document}